\documentclass[preprints,article,accept,moreauthors,pdftex]{Definitions/mdpi} 
\firstpage{1} 
\pubvolume{1}
\issuenum{1}
\articlenumber{0}
\pubyear{2021}
\copyrightyear{2020}
\datereceived{} 
\dateaccepted{} 
\datepublished{} 
\hreflink{https://doi.org/} 

\usepackage{graphicx}
\usepackage{subcaption}
\usepackage{mathtools}

\usepackage{algorithmic,algorithm}

\DeclareMathOperator{\EX}{\mathbb{E}}

\Title{Conditional Deep Gaussian Processes: multi-fidelity kernel learning}
\TitleCitation{Conditional DGP: multi-fidelity kernel learning}

\Author{Chi-Ken Lu $^{1,*}$\orcidA{} and Patrick Shafto $^{1,2}$}

\AuthorNames{Chi-Ken Lu and Patrick Shafto}

\AuthorCitation{Lu, C.-K.; Shafto, P.}

\address{%
$^{1}$ \quad Mathematics and Computer Science, Rutgers University, Newark, NJ 07102 \\
$^{2}$ \quad School of Mathematics, Institute for Advanced Study, Princeton 
}

\corres{Correspondence: cl1178@rutgers.edu
}

\abstract{ Deep Gaussian Processes (DGPs) were proposed as an expressive Bayesian model capable of a mathematically grounded estimation of uncertainty. The expressivity of DPGs results from not only the compositional character but the distribution propagation within the hierarchy. Recently, \cite{cutajar2019deep} pointed out that the hierarchical structure of DGP well suited modeling the multi-fidelity regression, in which one is provided sparse observations with high precision and plenty of low fidelity observations. We propose the conditional DGP model in which the latent GPs are directly supported by the fixed lower fidelity data. Then the moment matching method in~\cite{lu2020interpretable} is applied to approximate the marginal prior of conditional DGP with a GP. The obtained effective kernels are implicit functions of the lower-fidelity data, manifesting the expressivity contributed by distribution propagation within the hierarchy. The hyperparameters are learned via optimizing the approximate marginal likelihood. 
Experiments with synthetic and high dimensional data show comparable performance against other multi-fidelity regression methods, variational inference, and multi-output GP. We conclude that, with the low fidelity data and the hierarchical DGP structure, the effective kernel encodes the inductive bias for true function 
allowing the compositional freedom discussed in~\cite{havasi2018inference,ustyuzhaninov2020compositional}.     
}


\begin{document}

\section{Introduction}
Multi-fidelity regression refers to a category of learning tasks in which a set of sparse data is given to infer the underlying function but a larger amount of less precise or noisy observations is also provided. Multifidelity tasks frequently occur in various fields of science because precise measurement is often costly while approximate measurements are more affordable (see~\cite{kennedy2000predicting} and~\cite{wang2021multi} for example). The assumption that the more precise function is a function of the less precise one~\cite{kennedy2000predicting,cutajar2019deep} is shared in some hierarchical learning algorithms (e.g. one-shot learning in~\cite{salakhutdinov2012one}, meta learning~\cite{finn2017model}, and continual learning~\cite{titsias2019functional}). Thus, one can view the plentiful low fidelity data as a source of prior knowledge so the function can be efficiently learned with sparse data.

In Gaussian Process (GP) regression~\cite{rasmussen2006gaussian} domain experts can encode their knowledge into the combinations of covariance functions~\cite{duvenaud2013structure,sun2018differentiable}, building an expressive learning model. 
However, construction of an appropriate kernel becomes less clear when building a prior for the precise function in the context of multi-fidelity regression because the uncertainty, both epistemic and aleatoric, in the low fidelity function prior learned by the plentiful data should be taken into account. It is desirable to fuse the low fidelity data to an {\it effective} kernel as a prior, taking advantage of marginal likelihood being able to avoid overfitting, and then perform the GP regression as if only the sparse precise observations are given.

Deep Gaussian Process (DGP)~\cite{damianou2013deep} and the similar models~\cite{snelson2004warped,lazaro2012bayesian} are expressive models with hierarchical composition of GPs. As pointed out in~\cite{cutajar2019deep}, hierarchical structure is particularly suitable for fusing data of different fidelity into one learning model. Although full Bayesian inference is promising in obtaining expressiveness while avoiding overfitting, exact inference is not tractable and approximate solutions such as the variational approach~\cite{salimbeni2017doubly,salimbeni2019deep,haibin2019implicit} are employed. Ironically, the major difficulty in inference comes from marginalization of the the latent GPs in Bayesian learning, which, on the flip side, is also why overfitting can be prevented. 

We propose a conditional DGP model in which the intermediate GPs are supported by the lower fidelity data. We also define the corresponding marginal prior distribution which is obtained by marginalizing all GPs except the exposed one. For some families of kernel compositions, we previously developed analytical method in calculating exact covariance in the marginal prior~\cite{lu2020interpretable}. As such, the method is applied here so the marginal prior is approximated as a GP prior with an effective kernel. The high fidelity data are then connected to the exposed GP, and the hyperaparameters throughout the hierarchy are optimized via the marginal likelihood. 
Our model therefore captures the expressiveness embedded in hierarchical composition, retains the Bayesian character hinted in the marginal prior, but loses the non-Gaussian aspect of DGP. From the analytical expressions, one can partially understand the propagation of uncertainty in latent GPs as it is responsible for the non-stationary aspect of effective kernels. Moreover, the compositional freedom
\footnote{It refers to the fact that there are different compositions $f_2(f_1(x)))$ which may result in the same function $f(x)$. For instance $\sin2x$ can be decomposed into $f_1(x)=2x$ and $f_2(x)=\sin x$, and $f_1(x)=\sin x$ and $f_2(x)=2x\sqrt{1-x^2}$.} 
in a DGP model~\cite{havasi2018inference,ustyuzhaninov2020compositional} can be shown to be intact in our approach.    

The paper is organized as follows. In Sec. 2, we review the literature of multi-fidelity regression model and deep kernels. A background of GP, DGP, and the moment matching method is introduced in Sec. 3. The conditional DGP model defined as a marginal prior and the exact covariance associated with two family of kernel compositions are discussed in Sec. 4. The method of hyperparameter learning is given in Sec. 5, and the simulation of synthetic and high dimensional multi-fidelity regression in a variety of situations are presented in Sec. 6. A brief discussion followed by the conclusion appear in Sec. 7 and 8, respectively.

\section{Related work}
Assuming autoregressive relations between data of different fidelity, \citet{kennedy2000predicting} proposed the AR1 model for multi-fidelity regression tasks. \citet{le2014recursive} improved computational efficiency with a recursive multi-fidelity model. Deep-MF~\citep{raissi2016deep} mapped the input space to the latent space and followed the work in~\citet{kennedy2000predicting}. NARGP~\citep{perdikaris2017nonlinear} stacked a sequence of GPs in which the posterior mean about the low-fidelity function is passed to the input of the next GP while the associated uncertainty is not. GPAR~\citep{requeima2019gaussian} uses similar conditional structure between functions of interest. MF-DGP in~\citep{cutajar2019deep} exploited the DGP structure for the multi-fidelity regression tasks and used the approximate variational inference in~\citep{salimbeni2017doubly}. Multi-output GPs~\citep{alvarez2011kernels,parra2017spectral} regard the observations from different data sets as realization of vector-valued function; \cite{bruinsma2020scalable} modeled the multi-output GP using general relation between multiple target functions and multiple hidden functions. Alignment learning~\citep{kaiser2018bayesian,kazlauskaite2019gaussian} is application of warped GP~\citep{snelson2004warped,lazaro2012bayesian} to time series data. We model the multi-fidelity regression as a kernel learning, effectively taking the space of functions representing the low fidelity data into account. 

As for general studies of deep and non-stationary kernels, \citet{williams1997computing} and \citet{cho2009kernel} used the basis of error functions and Heaviside polynomial functions to obtain the arc-sine and arc-cosine kernel functions, respectively, of neural networks. \citet{duvenaud2014avoiding} employed the analogy between neural network and GP, and constructed the deep kernel for DGP. \citet{dunlop2018deep} analyzed variety of non-stationary kernel compositions in DGP, and \citet{shen2020learning} provided an insight from Wigner transformation of general two-input functions.
\citet{wilson2016deep} proposed the general recipe for constructing the deep kernel with neural networks. \citet{Daniely2016TowardDU} computed the deep kernel from the perspective of two correlated random variables. \citet{mairal2014convolutional} and \citet{van2017convolutional} studied the deep kernels in the convolutional models. The moment matching method~\cite{lu2020interpretable} allows to obtain the effective kernel encoding the uncertainty in learning the lower fidelity function. 



\section{Background}

\subsection{Gaussian Process and Deep Gaussian Process}

Gaussian Process (GP)~\cite{rasmussen2006gaussian} is a popular Bayesian learning model in which the prior over a continuous function is modeled as a Gaussian. Denoted by $f\sim\mathcal{GP}(\mu,k)$, the collection of any finite function values $f({\bf x}_{1:N})$ with ${\bf x}\in\mathbb R^d$ has the mean $\EX[f({\bf x}_i)]=\mu({\bf x}_i)$ and covariance $\EX\{[f({\bf x}_i)-\mu({\bf x}_i)][f({\bf x}_j)-\mu({\bf x}_j)]\}=k({\bf x}_i, {\bf x}_j)$. Thus a continuous and deterministic mean function $\mu(\cdot)$ and a positive definite kernel function $k(\cdot,\cdot)$ fully specify the stochastic process. It is common to consider the zero-mean case and write down the prior distribution, $p({\bf f|X})=\mathcal N(0,K)$ with covariance matrix $K$. In the setting of a regression task with input and output of data $\{{\bf X, y}\}$, the hyperparameters in the mean and kernel functions can be learned by optimizing the marginal likelihood, $p({\bf y|X})=\int d{\bf f}p({\bf y|f})p({\bf f|X})$.

Deep Gaussian Process (DGP) was proposed in~\cite{damianou2013deep} as a hierarchical composition of GPs for superior expressivity. From a generative view, the distribution over the composite function $f({\bf x})=f_L\circ f_{L-1}\circ\cdots\circ f_1({\bf x})$ is a serial product of Gaussian conditional distribution, 
\begin{equation}
	p({\bf F}_L,{\bf F}_{L-1},\cdots,{\bf F}_1|{\bf x})=p({\bf F}_L|{\bf F}_{L-1})p({\bf F}_{L-1}|{\bf F}_{L-2})\cdots p({\bf F}_1|{\bf X})\:,
\end{equation} in which the capital bolded face symbol ${\bf F}_i$ stands for a multi-output GP in $i$th layer and $p({\bf F}_i|{\bf F}_{i-1})=\mathcal N(0,K({\bf F}_{i-1}.{\bf F}_{i-1}))$. The above is the general DGP, and the width in each layer is denoted by $H_i:=|{\bf F}_i|$. In such notation, the zeroth layer represents the collection of inputs ${\bf X}$. Here, we shall consider the DGP with $L=2$ and $H_2=H_1=1$ and the three-layer counterpart.

The intractability of exact inference is a result of the fact that the random variables ${\bf F}_i$ for $L>i>0$ appear in the covariance matrix $K$. In a full Bayesian inference, the random variables are marginalized in order to estimate the evidence $p({\bf y|X})$ associated with the data. 

\subsection{Multi-fidelity Deep Gaussian Process}

The multi-fidelity model in~\cite{kennedy2000predicting} considered the regression task for a data set consisting of observations measured with both high and low precision. For simplicity, the more precise observations are denoted by $\{{\bf X, y}\}$ and those with low precision by $\{{\bf X}_1,{\bf y}_1\}$. The main assumption made in~\cite{kennedy2000predicting} is that the less precise observations shall come from a function $f_1({\bf x})$ modeled by a GP with zero mean and kernel $k$, while the more precise ones come from the combination $f({\bf x})=\alpha f_1({\bf x})+h({\bf x})$. With the residual function $h$ being a GP with kernel $k_h$, one can jointly model the two subsets with the covariance within precise observations $\EX[f({\bf x}_i)f({\bf x}_j)]=\alpha^2k_{ij}+k_{h_{ij}}$, within the less precise ones $\EX[f_1({\bf x}_i)f_1({\bf x}_j)]=k_{ij}$, and the mutual covariance $\EX[f({\bf x}_i)f_1({\bf x}_j)]=\alpha k_{ij}$. $k_{ij}$ refers to the covariance between the two inputs ${\bf x}_i$ and ${\bf x}_j$. 

The work in~\cite{cutajar2019deep} generalized the above the linear relationship between the more and less precise functions to a nonlinear one, i.e. $f({\bf x})=f_2(f_1({\bf x}))+{\rm noise}$. The hierarchical structure in DGP is suitable for the nonlinear modeling. The variational inference scheme~\cite{salimbeni2017doubly} was employed to evaluate the evidence lower bounds (ELBOs) for the data with all levels of precision, and the sum over all ELBOs is the objective for learning the hyperparameters and inducing points.

\subsection{Covariance in marginal prior of DGP}
The variational inference, e.g.~\cite{salimbeni2017doubly}, starts with connecting the joint distribution $p(f_1,f_2|{\bf X})$ with data ${\bf y}$, followed by applying the Jensen's inequality along with an approximate posterior in evaluating the ELBO. In contrast, we proposed in~\cite{lu2020interpretable} that the marginal prior for the DGP,
\begin{equation}
	p({\bf f}|{\bf X}) = \int d{\bf f}_1 p({\bf f}_2|{\bf f}_1)p({\bf f}_1|{\bf X})\:,\label{true_marginalprior} 
\end{equation} with the bolded face symbols representing the set of function values, $f(\cdot)=f_2(f_1(\cdot))$, $f_2(\cdot)$, and $f_1(\cdot)$, can be approximated as a GP, i.e. $q({\bf f}|{\bf X})=\mathcal N(0,K_{\rm eff})$ in the zero-mean case. The matching of covariance in $p$ and $q$ leads to the closed form of effective covariance function for certain families of kernel compositions. The SE[SC] composition, i.e. the squared exponential and squared cosine kernels being used in the GPs for $f_2|f_1$ and $f_1$, respectively, is an example. With the intractable marginalization over the intermediate $f_1$ being taken care of in the moment matching approximation, one can evaluate the approximate marginal likelihood for the data set $\{{\bf X, y}\}$,
\begin{equation}
	p({\bf y|X})\approx\int d{\bf f}p({\bf y|f})q({\bf f|X})\:.
\end{equation} In the following, we shall develop along the line of~\cite{lu2020interpretable} the approximate inference for the multi-fidelity data consisting of precise observations $\{{\bf X,y}\}$ and less precise observations $\{{\bf X}_{1:L-1},{\bf y}_{1:L-1}\}$ with the $L$-layer width-1 DGP models. The effective kernels $k_{\rm eff}$ shall encode the knowledge built on these less precise data, which allows modeling the precise function even with a sparse data set.

\section{Conditional DGP and multi-fidelity kernel learning}

In the simplest case, we are given two subsets of data, $\{{\bf X, y}\}$ with high precision and $\{{\bf X}_1, {\bf y}_1\}$ with low precision. We can start with defining the conditional DGP in terms of the marginal prior,
\begin{equation}
	p({\bf f|X,X}_1,{\bf y}_1)=\int d{\bf f}_1 p({\bf f}_2|{\bf f}_1)p({\bf f}_1|{\bf X, X}_1,{\bf y}_1)\:,\label{marginal_prior}
\end{equation} where the Gaussian distribution $p({\bf f}_1|{\bf X, X}_1,{\bf y}_1)=\mathcal N(f_1({\bf x}_{1:N})|{\bf m},\Sigma)$ has the conditional mean in the vector form,
\begin{equation}
	{\bf m}=K_{{\bf X,X}_1}K_{{\bf X}_1,{\bf X}_1}^{-1}{\bf y}_1\:,\label{cond_mean}
\end{equation} and the conditional covariance in the matrix form,
\begin{equation}
	\Sigma = K_{{\bf X, X}} - K_{{\bf X,X}_1}K_{{\bf X}_1,{\bf X}_1}^{-1}K_{{\bf X}_1,{\bf X}}\:.\label{cond_cov}
\end{equation} The matrix $K_{{\bf X,X}_1}$ registers the covariance among the inputs in ${\bf X}$ and ${\bf X}_1$, and likewise for $K_{{\bf X,X}}$ and $K_{{\bf X}_1,{\bf X}_1}$. Thus the set of function values $f_1({\bf x}_{1:N})$ associated with the $N$ inputs in ${\bf X}$ are supported by the low fidelity data. 

The data $\{{\bf X, y}\}$ with high precision are then associated with the function $f({\bf x})=f_2(f_1({\bf x}))$. Following the previous discussion, we may write down the true evidence for the precise data conditioned on the less precise ones shown below, 
\begin{equation}
	p({\bf y|X,X}_1,{\bf y}_1)=\int d{\bf f}p({\bf y|f})p({\bf f|X,X}_1,{\bf y}_1)=\int d{\bf f}_1d{\bf f}_2p({\bf y|f}_2)
	p({\bf f}_2|{\bf f}_1)p({\bf f}_1|{\bf X, X}_1,{\bf y}_1)\:.\label{true_evidence}
\end{equation} To proceed with the moment matching approximation of the true evidence in Eq.~(\ref{true_evidence}), one needs to find the effective kernel in the approximate distribution $q({\bf f|X,X}_1,{\bf y}_1)=\mathcal N(0,K_{\rm eff})$ and replace the true distribution in Eq.~(\ref{marginal_prior}) with the approximate distribution, 
\begin{equation}
	p({\bf y|X,X}_1,{\bf y}_1)\approx\int d{\bf f}p({\bf y|f})q({\bf f|X,X}_1,{\bf y}_1)=\mathcal N({\bf y}|0,K_{\rm eff}+\sigma_n^2I_N)\:.\label{approx_evidence}
\end{equation} 

Therefore, the learning in the conditional DGP includes the hyperparameters in the exposed GP, $f_2|f_1$, and those in the intermediate GP, $f_1$. Standard gradient descent is applied to above approximate marginal likelihood. One can customize the kernel $K_{\rm eff}$ in the GPy~\cite{gpy2014} framework and implement the gradient components $\partial K_{\rm eff}/\partial\theta$ with $\theta\in\{\sigma_{1,2},\ell_{1,2}\}$ in the optimization.

\subsection{Effective kernels for SE[~] and SC[~] compositions}
Here, we consider the conditional DGP with two-layer and width-1 hierarchy, and focus on the cases where the exposed GP for $f_2|f_1$ in Eq.~(\ref{marginal_prior}) uses the squared exponential (SE) kernel or the squared cosine (SC) kernel. Following~\cite{lu2020interpretable}, the exact covariance in the marginal prior Eq.~(\ref{marginal_prior}) is calculated,
\begin{equation} 
	\EX_{{\bf f}}[f({\bf x}_i)f({\bf x}_j)]:=\EX_{{\bf f}_1}[\EX_{{\bf f}_2|{\bf f}_1}[f_2(f_1({\bf x}_i))f_2(f_1({\bf x}_j))]]=\int d{\bf f}_1\ k_2(f_1({\bf x}_i),f_1({\bf x}_j))p({\bf f}_1|{\bf X, X}_1, {\bf y}_1)\:.
\end{equation} Thus, when the exposed GP has the kernel $k_2$ in the exponential family, the above integral is tractable and the analytic $k_{\rm eff}$ can be implemented as a customized kernel. The following two lemmas from~\cite{lu2020interpretable} are useful for the present case with a nonzero conditional mean and a conditional covariance in $f_1$.

\begin{Lemma} (Lemma 2 in~\cite{lu2020interpretable}) For a vector of Gaussian random variables $g_{1:n}\sim\mathcal N({\bf m},{\bf C})$, the expectation of exponential quadratic form $\exp[-\frac{1}{2}Q(g_1,g_2,\cdots,g_n)]$ with $Q(g_{1:n})=\sum_{i,j}A_{ij}g_ig_j\geq0$ has the following closed form,
\begin{equation}
    \EX[e^{-\frac{1}{2}Q(g_{1:n})}] =
    \frac{\exp\left[-\frac{1}{2}{\bf m}^T{\bf C}^{-1}(I_n-(I_n+{\bf CA})^{-1}){\bf m}\right]}
    {\sqrt{|I_n+{\bf C}{\bf A}|}}\:.\label{se_composition}
\end{equation} The $n$-dimensional matrix ${\bf A}$ appearing in the quadratic form $Q$ is symmetric.
\end{Lemma}

\begin{Lemma} (Lemma 3 in~\cite{lu2020interpretable})
With the same Gaussian vector ${\bf g}$ in Theorem~1,  the expectation value of the exponential inner product $\exp[{\bf a}^t{\bf g}]$ between ${\bf g}$ and a constant vector ${\bf a}$ reads,
\begin{equation}
    \EX[e^{{\bf a}^t{\bf g}}]=\exp\{{{\bf a}^t{\bf m}+\frac{1}{2}Tr[{\bf C}{\bf a}{\bf a}^t]}\}\:,\label{sc_composition}
\end{equation} where the transpose operation on column vector is denoted by the superscript.
\end{Lemma}
We shall emphasize that our previous work~\cite{lu2020interpretable} did not discuss the cases when the intermediate GP for $f_1$ is conditioned on the low precision data $\{{\bf X}_1,{\bf y}_1\}$. Thus the conditional mean in $f_1$ [Eq.~(\ref{cond_mean})] as well as the non-stationary covariance [Eq.~(\ref{cond_cov})] were not considered in~\cite{lu2020interpretable}.

\begin{Lemma} The covariance in the marginal prior with a SE $k_2(x,y)=\sigma_2^2\exp[-(x-y)^2/2\ell_2^2]$ in the exposed GP can be calculated analytically. With the Gaussian conditional distribution, $p({\bf f}_1|{\bf X, X}_1,{\bf y}_1)$, supported by the low fidelity data, the effective kernel reads, 
\begin{equation}
    k_{\rm eff}({\bf x}_i,{\bf x}_j)
    =\frac{\sigma_2^2}
    {\sqrt{1+\delta_{ij}^2/\ell_2^2}}\exp\left[-\frac{(m_i-m_j)^2}{2(\ell_2^2+\delta_{ij}^2)}
    \right]\:,\label{kernel:se}
\end{equation} where $m_{i,j}:=\EX[f_1({\bf x}_{i,j})]$ and $c_{ij}:={\rm cov}(f_1({\bf x}_i),f_1({\bf x}_j))$ being the conditional mean and conditional covariance, respectively, of $p({\bf f}_1|{\bf X, X}_1,{\bf y}_1)$ at the inputs ${\bf x}_{i,j}$. The positive parameter $\delta_{ij}^2:=c_{ii}+c_{jj}-2c_{ij}$ and the the length scale $\ell_2$ in $k_2$ dictates how the uncertainty in $f_1({\bf x})$ affects the function composition. 
\end{Lemma}

\begin{proof}
For SE[~] composition, one can represent the kernel $k_2=\exp\{-[f_1({\bf x}_i)-f_1({\bf x}_j)]^2/2\}$ as an exponential quadratic form $\exp[-\frac{Q}{2}]$ with $Q={\bf f}_1^t{\bf A}{\bf f}_1$ with ${\bf A}=\begin{psmallmatrix}1 & -1 \\-1 &1\end{psmallmatrix}$. $\ell_2=1$ is set for ease of notation. Now ${\bf f}_1$ is a bivariate Gaussian variable with mean ${\bf m}$ and covariance matrix ${\bf C}$. To calculate the expectation value in Eq.~(\ref{kernel:se}), we need to compute the following 2-by-2 matrix and one can show $\left[I_2+{\bf CA}\right]^{-1}$ can be reduced to
\begin{equation}
    \frac{1}{1+c_{ii}+c_{jj}-2c_{ij}}\left(
    \begin{matrix}1+c_{jj}-c_{ij} & c_{ii}-c_{ij} \\c_{jj}-c_{ij} & 1+c_{ii}-c_{ij}
    \end{matrix}
    \right)\:.
\end{equation} The seemingly complicated matrix in fact is reducible as one can show that $I_2-[I_2+{\bf CA}]^{-1}={\bf CA}/(1+\delta_{ij}^2)$. Thus, the exponent in right-hand side of Eq.~(\ref{se_composition}) reads $[-\frac{1}{2}{\bf m}^t{\bf A}{\bf m}/(1+\delta_{ij}^2)]$. Moreover, it is easy to show that the determinant $|I_2+{\bf CA}|=(1+\delta_{ij}^2)$. By restoring back the length scale $\ell_2$, the kernel in Eq.~(\ref{kernel:se}) is reproduced.
\end{proof}
A few observations are in order. First, we can rewrite  $\delta^2_{ij}=\begin{psmallmatrix}1 & -1 \end{psmallmatrix}\begin{psmallmatrix}c_{ii} & c_{ij} \\c_{ji} &c_{jj}\end{psmallmatrix}\begin{psmallmatrix}1\\-1\end{psmallmatrix}$, which guarantees the positiveness of $\delta^2_{ij}$ as the two-by-two sub-block of covariance matrix is positive-definite. 
Secondly,  
there are deterministic and probabilistic aspects of the kernel in Eq.~(\ref{kernel:se}). When the the length scale $\ell_2$ is very large, the term $\delta^2$ encoding the uncertainty in $f_1$ becomes irrelevant and the kernel is approximately a SE kernel with the input transformed via the conditonal mean [Eq.~(\ref{cond_mean})], which is reminiscent of the deep kernel proposed in~\cite{wilson2016deep} where GP is stacked on the output of a DNN. The kernel used in~\cite{perdikaris2017nonlinear} similarly considers the conditional mean in $f_1$ as a deterministic transformation while the uncertainty is ignored. 
On the other hand, when $\delta^2$ and $\ell_2^2$ are comparable, it means that
the (epistemic) uncertainty in $f_1$ shaped by the supports ${\bf y}_1$ is relevant. The effective kernel then represents the covariance in the ensemble of GPs, each of which receives the inputs transformed by one $f_1$ sampled from the intermediate GP. 
Thirdly, we shall stress that the appearance of $\delta^2$ is a signature of marginalization over the latent function in deep probabilistic models. Similar square distance also appeared in \cite{duvenaud2014avoiding} where the effectively deep kernel was derived from a recursive inner product in the space defined by neural network feature functions.  

In the following lemma, we consider the composition where the kernel in outer layer is squared cosine, $k_h(x,y)=(\sigma_2^2/2)\{1+\cos[(x-y)/\ell_2]\}$, which is a special case of spectrum mixture kernel~\cite{wilson2013gaussian}. 
\begin{Lemma}
The covariance in $f$ of the marginal prior with SC kernel used in the exposed GP is given below, 
\begin{equation}
    k_{\rm eff}({\bf x}_i,{\bf x}_j)
    =\EX_p[f_if_j]
    =\frac{\sigma_2^2}{2}\left[1+
    \cos(m_i-m_j)
    \exp(-\frac{\delta_{ij}^2}{2\ell_2^2})\right]\:,\label{kernel:cos}
\end{equation} where  
$\delta^2_{ij}$ has been defined in the previous lemma.
\end{Lemma} 
The form of product of cosine and exponential kernels is similar with the deep spectral mixture kernel~\cite{wilson2016deep}. In our case the cosine function has the warped input $m(x_i)-m(x_j)$, but the exponential function has the input $c(x_i,x_i)+c(x_j,x_j)-2c(x_i,x_j)$ due to the conditional covariance in the intermediate GP. 


\subsection{Samples from the marginal prior}
\begin{figure}[ht]
\centering
\includegraphics[width=0.45\columnwidth]{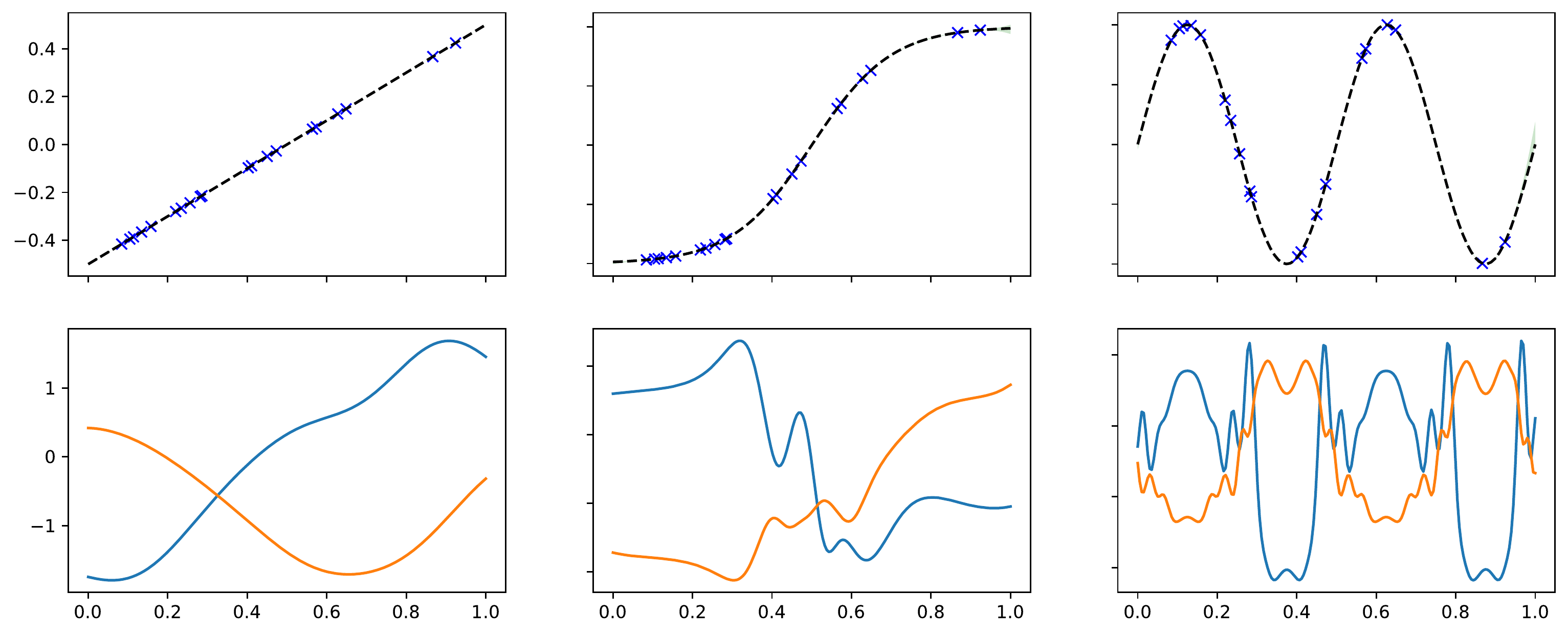}
\includegraphics[width=0.45\columnwidth]{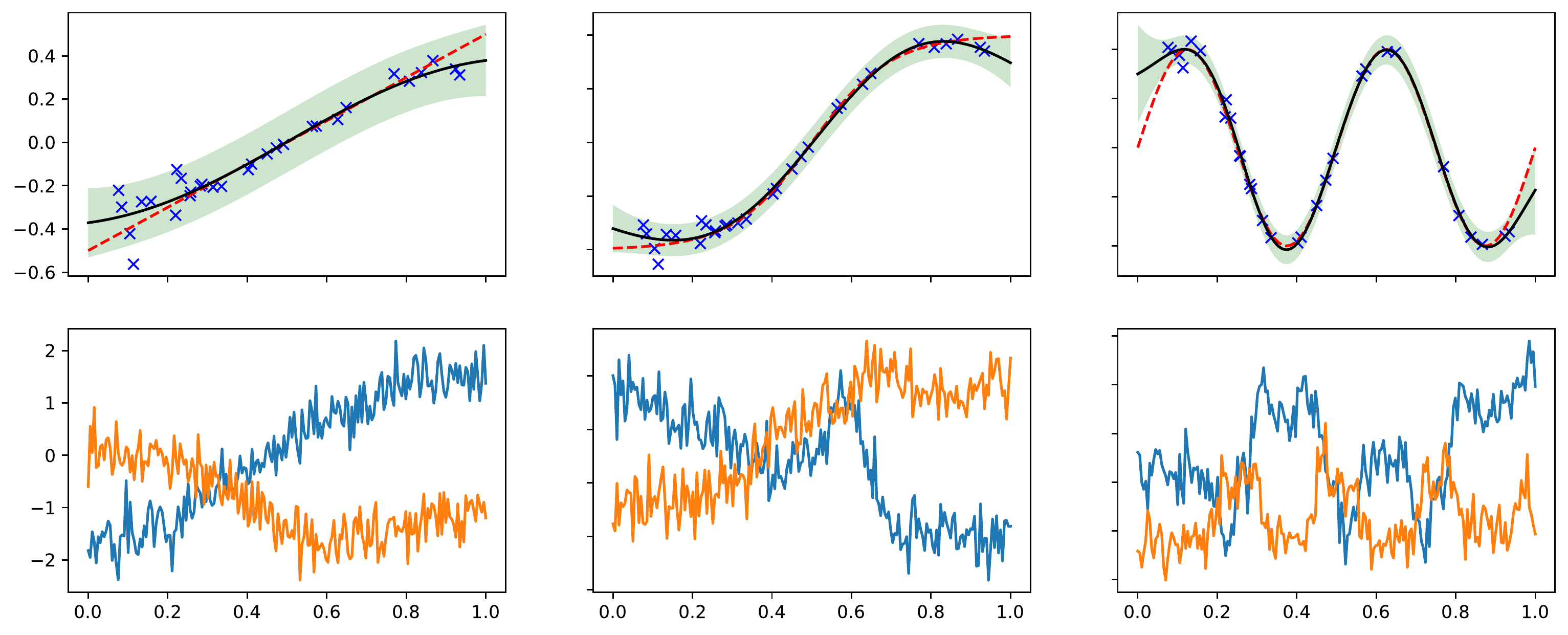}
\caption{
Sampling random functions from the approximate marginal prior $q({\bf f})$ which carries the effective kernel in Eq.~(\ref{kernel:se}). The low-fidelity data ${\bf X}_1,{\bf y}_1$, marked by the cross symbols, and the low fidelity function $f_1|{\bf X}_1,{\bf y}_1$ and the uncertainty are shown in the top row. 
First, second, and third columns: the uncertainty in ${\bf X}_1,{\bf y}_1$ is negligible so $f_1$ is nearly a deterministic function in top panels, so the effective kernels are basically kernels with warped input. The samples from $q$ are shown in the bottom panels.
Fourth, fifth, and sixth columns: uncertainty in ${\bf X}_1,{\bf y}_1$ (top) generates the samples from $q$ which 
carry additional high-frequency signals due to the non-stationary $\delta^2$ in Eq.~(\ref{kernel:se}).}\label{sample}
\end{figure}

Now we study the samples from the approximate marginal prior with the effective kernel in Eq.~(\ref{kernel:se}). We shall vary the low fidelity data ${\bf X}_1, {\bf y}_1$ to see how they affect the inductive bias for the target function. 
See the illustration in Fig.~\ref{sample}. The top row displays $f_1|{\bf X}_1, {\bf y}_1$, which is obtained by a standard GP regression. The conditional covariance and condition mean are then fed into the effective kernel in Eq.~(\ref{kernel:se}), and so we can sample functions from the prior carrying the effective kernel. The samples are displayed in the bottom row. The left three columns show the case where ${\bf y}_1$ are noiseless observations of some simple functions, while the right three panels show those with ${\bf y}_1$ from noisy observations of the same three functions.

In the clean cases (left three columns), it can be seen that $f_1|{\bf X}_1, {\bf y}_1$ is nearly a deterministic function (top panels) given the sufficient amount of observations in $\{{\bf X}_1,{\bf y}_1\}$. In fact, the left most panel is equivalent to the samples from a SE kernel as $f_1$ is the identity function. 
Moving to the second column, one can see the effect of nonlinear warping, $f_1$ generates additional kinks. The case on the third column with periodic warping results in periodic patterns to the sampled functions.

Next, we shall see the effect of uncertainty in $f_1|{\bf X}_1, {\bf y}_1$ (top panels) on the sampled functions (bottom panels).
As shown in the right three columns, the increased uncertainty (shown by shadow region) in $f_1$ has the effects: generating the weak and high frequency signal due to the non-stationary $\delta^2$ in Eq.~(\ref{kernel:se}). We stress that these weak signals are not white noise. The high noise in periodic $f_1$ in the 6th column even reverses the sign of sampled functions in comparing with the 3rd column. Consequently, the expressivity of the effective kernel gets a contribution from the uncertainty in learning the low fidelity functions. 

\section{Method}

Since we approximate the marginal prior for the conditional DGP with a GP, the corresponding approximate marginal likelihood should be the objective for jointly learning the hyperparameters including those in the exposed GP and the intermediate GPs. From the analytical expression for the effective kernel, e.g. Eq.~(\ref{kernel:se}), the gradient components include the explicit derivatives $\partial{K_{\rm eff}}/\partial{\sigma_2}$ and $\partial{K_{\rm eff}}/\partial{\ell_2}$ as well as those implicit derivatives which can be computed via chain rule, 
\begin{equation*}
	\frac{\partial K_{\rm eff}}{\partial\sigma_1}=\frac{\partial{K_{\rm eff}}}{\partial{\delta^2_{ij}}}\frac{\partial{\delta^2_{ij}}}{\partial \sigma_1}\:, 
\end{equation*} and 
\begin{equation*}
	\frac{\partial{K_{\rm eff}}}{\partial{\ell_1}}=\frac{\partial{K_{\rm eff}}}{\partial{\delta^2_{ij}}}\frac{\partial{\delta^2_{ij}}}{\partial \ell_1}+
	\frac{\partial{K_{\rm eff}}}{\partial{(m_i-m_j)}}\frac{\partial{(m_i-m_j)}}{\partial \ell_1}\:.
\end{equation*}

\begin{algorithm}[tb]
   \caption{A learning algorithm for conditional DGP multi-fidelity regression}
   \label{alg}
\begin{algorithmic}
   \STATE {\bfseries Input:} two sources of data, low-fidelity data $({\bf X}_1, {\bf y}_1)$ and high-fidelity data $({\bf X}, {\bf y})$, kernel $k_1$ for function $f_1$, and the test input ${\bf x}_*$.
   \STATE\ 1. $k_1$~=~Kernel(var=$\sigma_1^2$, lengthscale=$\ell_1$)
   ~\COMMENT{Initialize the kernel for inferring $g$}
   \STATE\ 2. model$_1$~=~Regression(kernel=$k_1$, data=${\bf X}_1$ and ${\bf y}_1$)~\COMMENT{Initialize regression model for $f_1$}
   \STATE\ 3. model$_1$.optimize()
   \STATE\ 4. ${\bf m}$, ${\bf C}$ = model$_1$.predict(input = ${\bf X,x}_*$, full-cov=true)~\COMMENT{Output pred. mean and post cov. of $f_1$}
   \STATE\ 5. $k_{\rm eff}$ = EffectiveKernel(var=$\sigma_2^2$, lengthscale=$\ell_2$, ${\bf m}$, ${\bf C}$)~\COMMENT{Initialize the effective kernel in Eq.~(\ref{kernel:se}) for SE[~] and Eq.~(\ref{kernel:cos}) for SC[~].}
   \STATE\ 6. model$_2$ = Regression(kernel=$k_{\rm eff}$, data = ${\bf X,y}$)~\COMMENT{Initialize regression model for $f$}
   \STATE\ 7. model$_2$.optimize()
   \STATE\ 8. $\mu_*$,$\sigma_*^2$ = model$_2$.predict(input=${\bf x}_*$)
   \STATE {\bfseries Output:} Optimal hyper-parameters $\{\sigma_{1,2}^2,\ell_{1,2}\}$ and predictive mean $\mu_*$ and variance $\sigma_*$ at ${\bf x}_*$.
\end{algorithmic}\label{algorithm}
\end{algorithm}

With the data consisting of observations of different fidelity, an alternative method can learn the hyperparameters associated with each layer of the hierarchy sequentially. See Algorithm~\ref{algorithm} for details. The low fidelity data are fed into the first GP regression model for inferring $f_1$ and the hyperparameters $\ell_1$ and $\sigma_1$. The conditional mean and conditional covariance in $f_1|{\bf X}_1,{\bf y}_1$ are then sent to the effective kernel. The second GP using the effective kernel is to infer the high fidelity function $f$ with the marginal likelihood for the high fidelity data being the objective. Optimization of the second model results in the hyperparameters $\ell_2$ and $\sigma_2$ in the second layer. Learning in the three-layer hierarchy can be generalized from the two-layer hierarchy. In the Appendix, a comparison of regression using the two methods is shown.  

\section{Results}
In this section, we shall present the results of multi-fidelity regression given low fidelity data ${\bf X}_1, {\bf y}_1$ and high fidelity ${\bf X}, {\bf y}$ and use the 2-layer conditional DGP model. The cases where there are three levels of fidelity can be generalized with the 3-layer counterpart. The toy demonstrations in Sec. 5.1 focus on data sets in which the target function is a composite, $f(x)=f_2(f_1(x))$. The low fidelity data are observations of $f_1(x)$ while the high fidelity are those of $f(x)$. The aspect of compositional freedom is discussed in Sec. 5.2, and the same target function shall be inferred with the same high fidelity data but the low fidelity data now result from a variety of functions. In Sec. 5.3, we switch to the case where the low fidelity data are also observations of the target function $f$ but with large noise. In Sec. 5.3, we compare our model with the work in~\cite{cutajar2019deep} on the data set with high dimensional inputs. 
 

\subsection{Synthetic two-fidelity function regression}

\begin{figure}[ht]
  \centering
  \includegraphics[width=0.3\columnwidth]{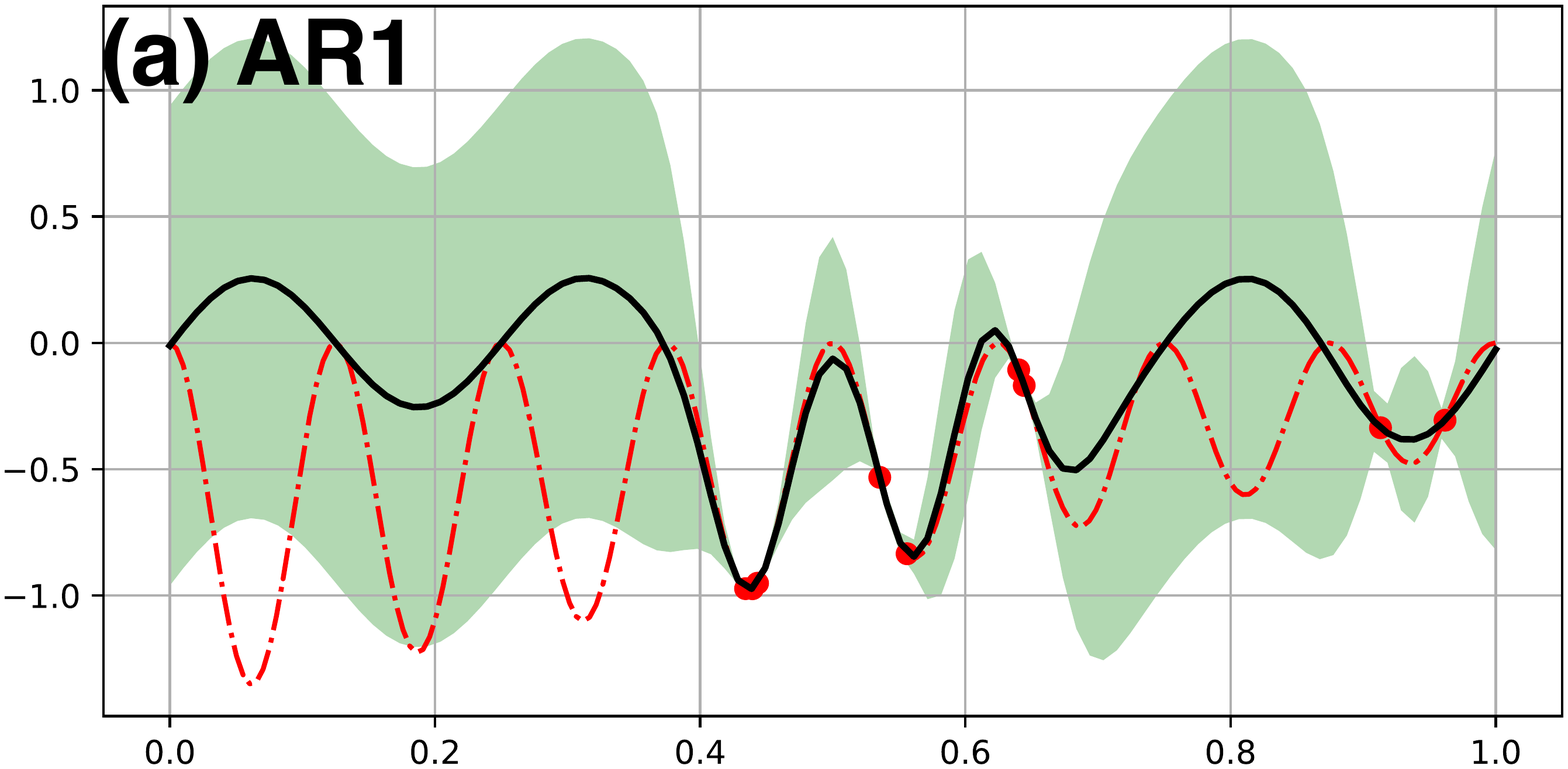}
  \includegraphics[width=0.3\columnwidth]{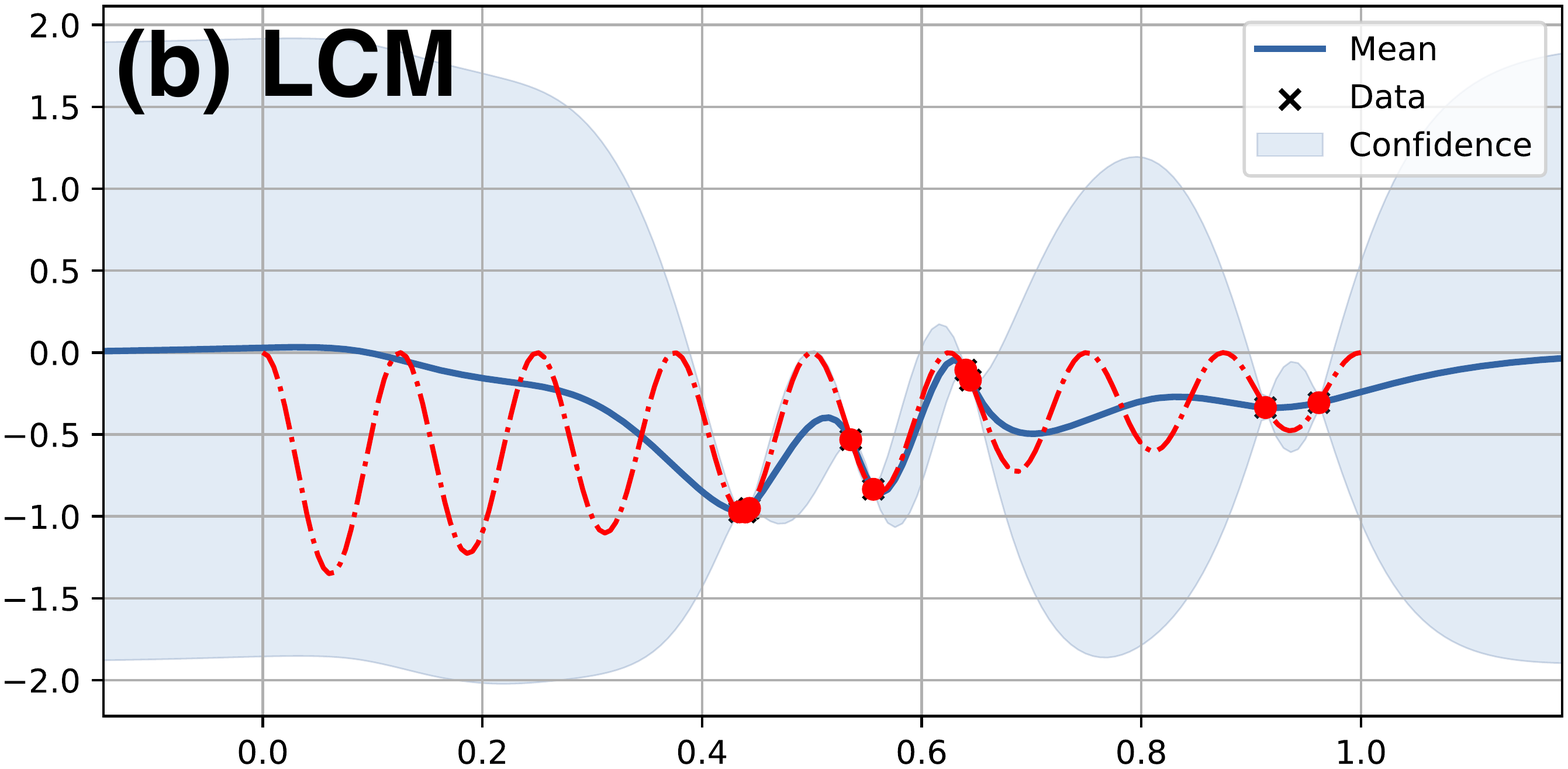}
  \includegraphics[width=0.3\columnwidth]{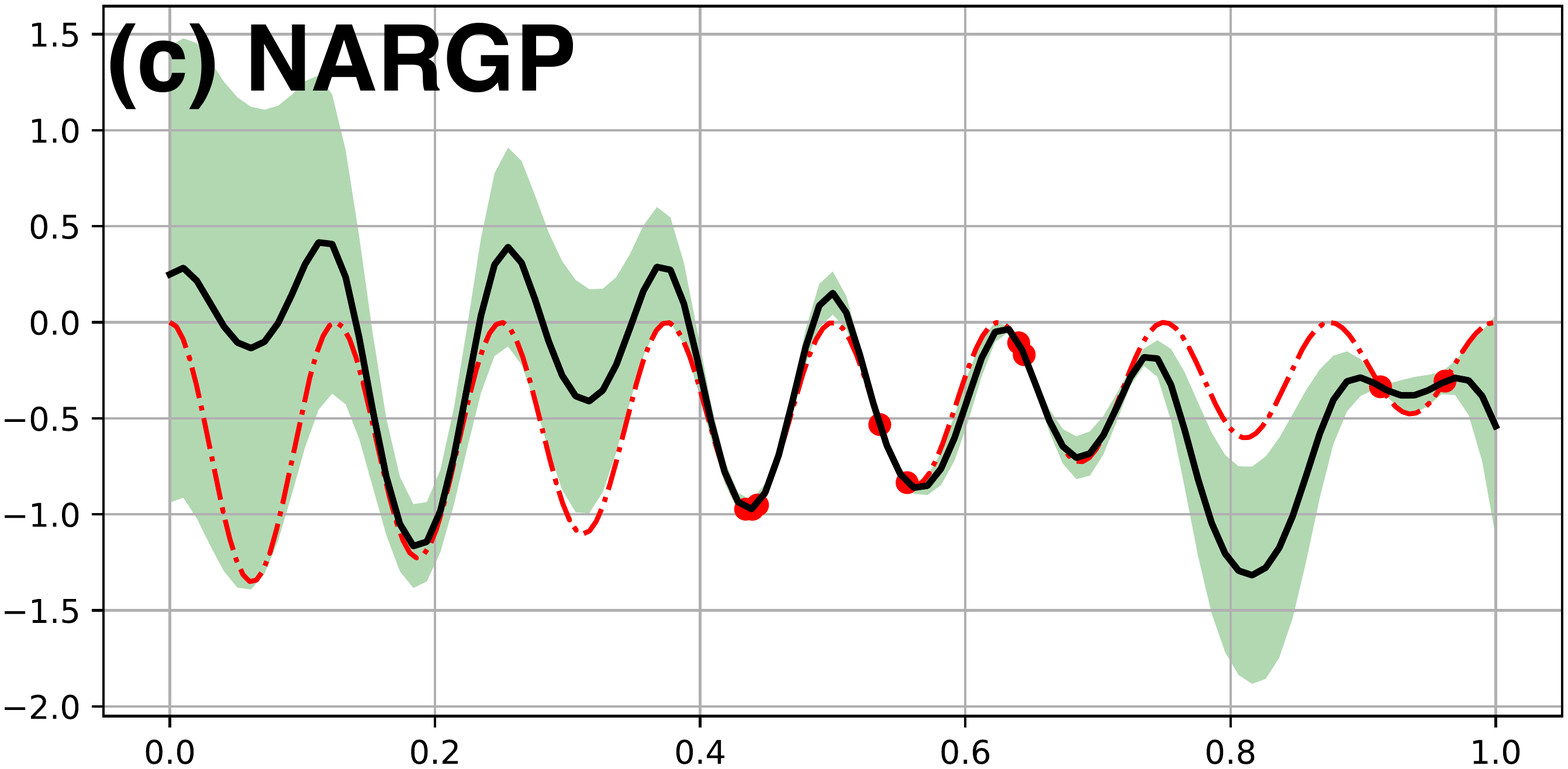}
  \includegraphics[width=0.3\columnwidth]{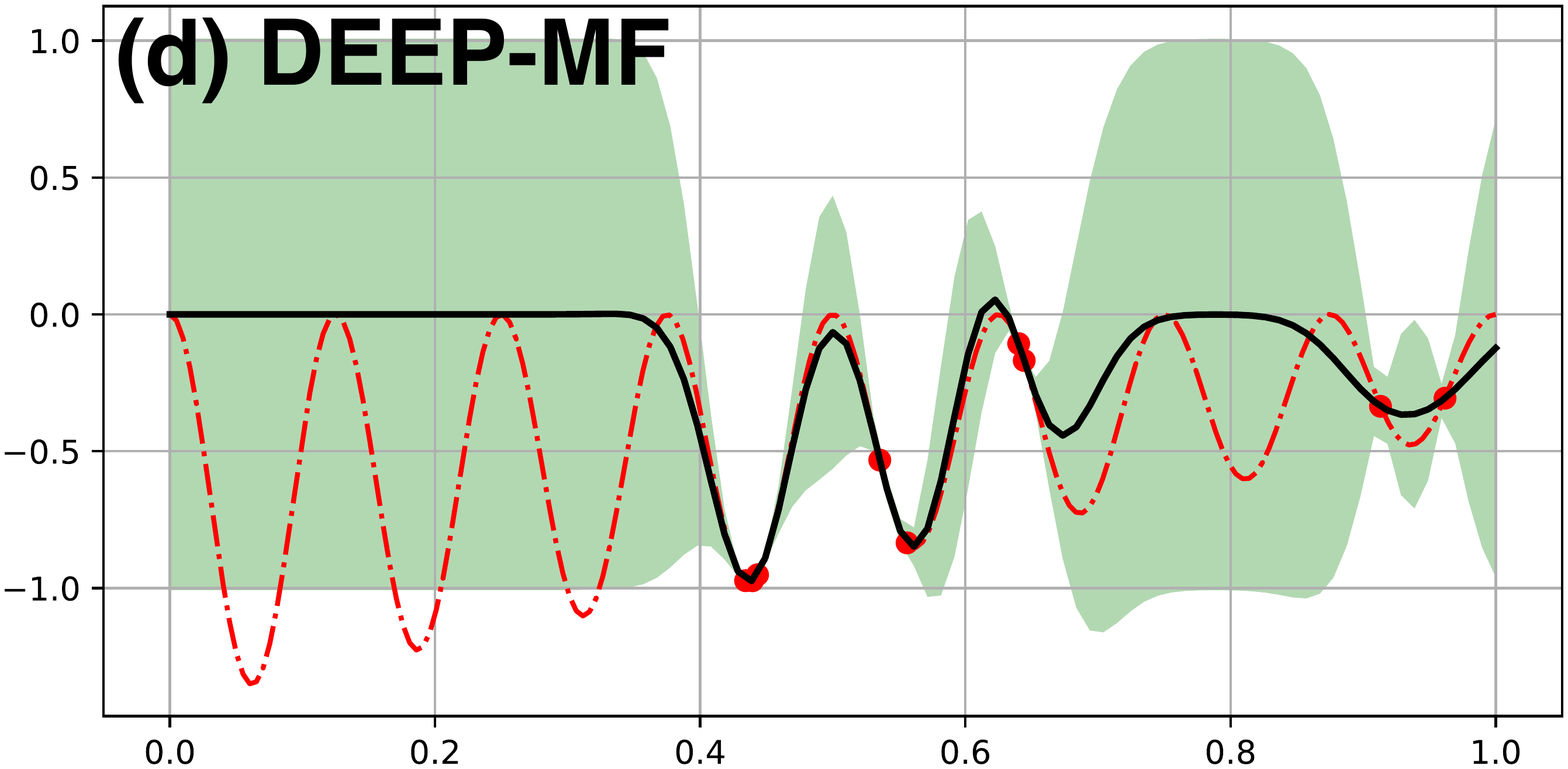}
  \includegraphics[width=0.3\columnwidth]{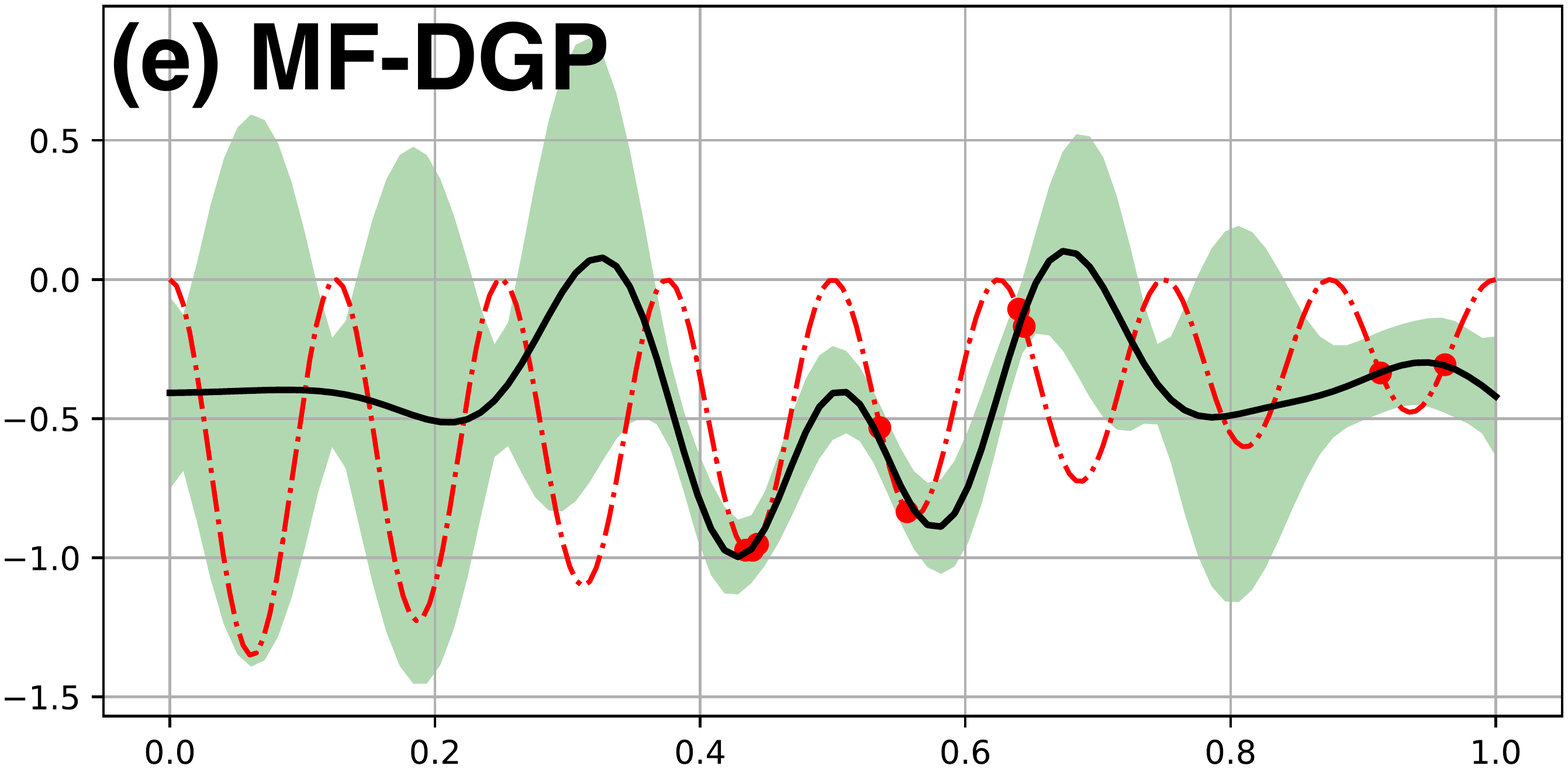}
  \includegraphics[width=0.3\columnwidth]{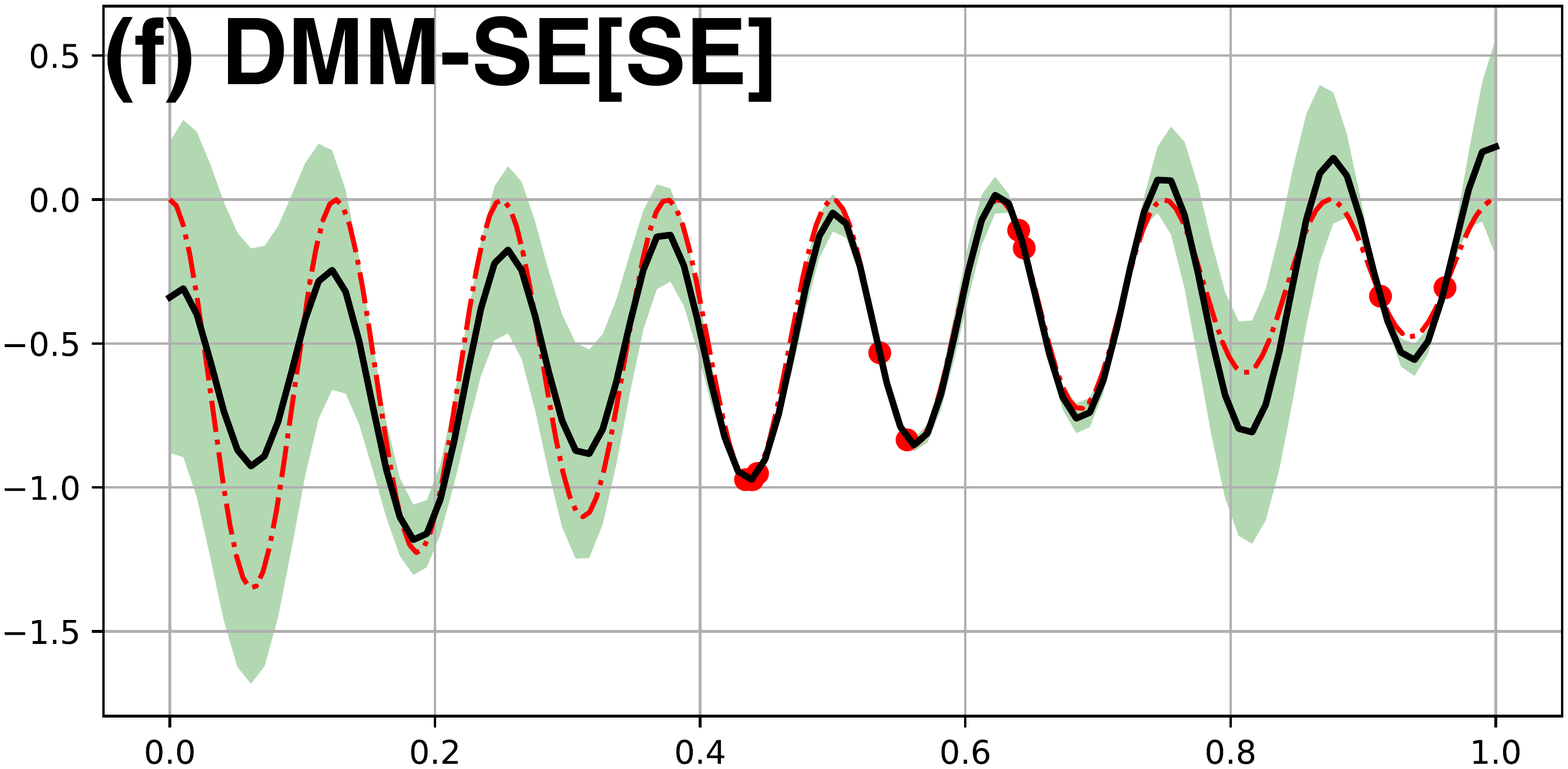}
  \caption{ 
  Multi-fidelity regression with 30 observations (not shown) of low fidelity $f_1(x)=\sin8\pi x$ and 10 observations (red dots) from the target function, $f(x)=(x-\sqrt{2})f_1^2(x)$ (shown in red dashed line). Only the target prediction (solid dark) and associated uncertainty (shaded) are shown: top row: (a) AR1, (b) LCM, (c) NARGP. bottom row: (d) DEEP-MF, (e) MF-DGP, (f) our model with SE[SE] kernel.
  }
  \label{exp_nonlinearA}
\end{figure}

The first example in Figure~\ref{exp_nonlinearA} consists of 10 random observations of the target function $f(x)=(x-\sqrt{2})f_1^2(x)$ (red dashed line) along with 30 observations of the low fidelity function $f_1(x)=\sin8\pi x$ (not shown). The 30 observations of $f_1$ with a period 0.25 in the range of $[0,1]$ is more than sufficient to reconstruct $f_1$ with high confidence. In contrast, the 10 observations of $f$ alone (shown in red dots) are difficult to reconstruct $f$ if a GP with SE kernel is used. The above figures demonstrate the results from a set of multi-source nonparametric regression methods which incorporate the learning of $f_1$ into the target regression of $f$.
Our result, the SE[SE] [panel (f)] kernel, and NARGP [panel (c)] successfully capture the periodic pattern inherited from the low fidelity function $f_1$, but the target function is fully covered in the confidence region in our prediction only. On the other hand, in the input space away from the target observations, AR1 [panel (a)] and MF-DGP [panel (e)] manages to only capture part of the oscillation. Predictions in LCM [panel(b)] and DEEP-MF [panel (d)] are reasonable near the target observations but fail to capture the oscillation away from these observations.

\begin{figure}[ht]
  \centering
  \includegraphics[width=0.3\columnwidth]{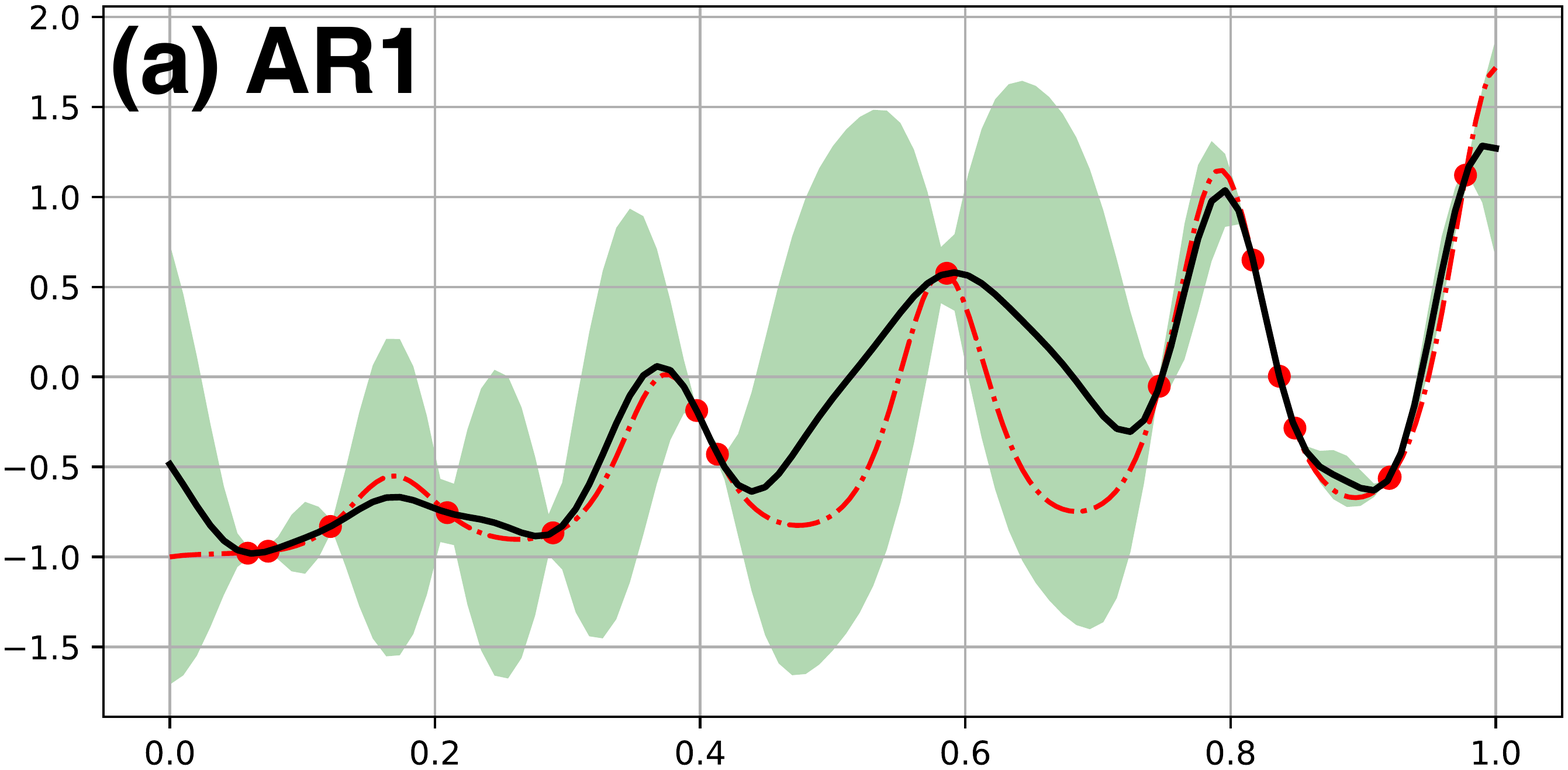}
  \includegraphics[width=0.3\columnwidth]{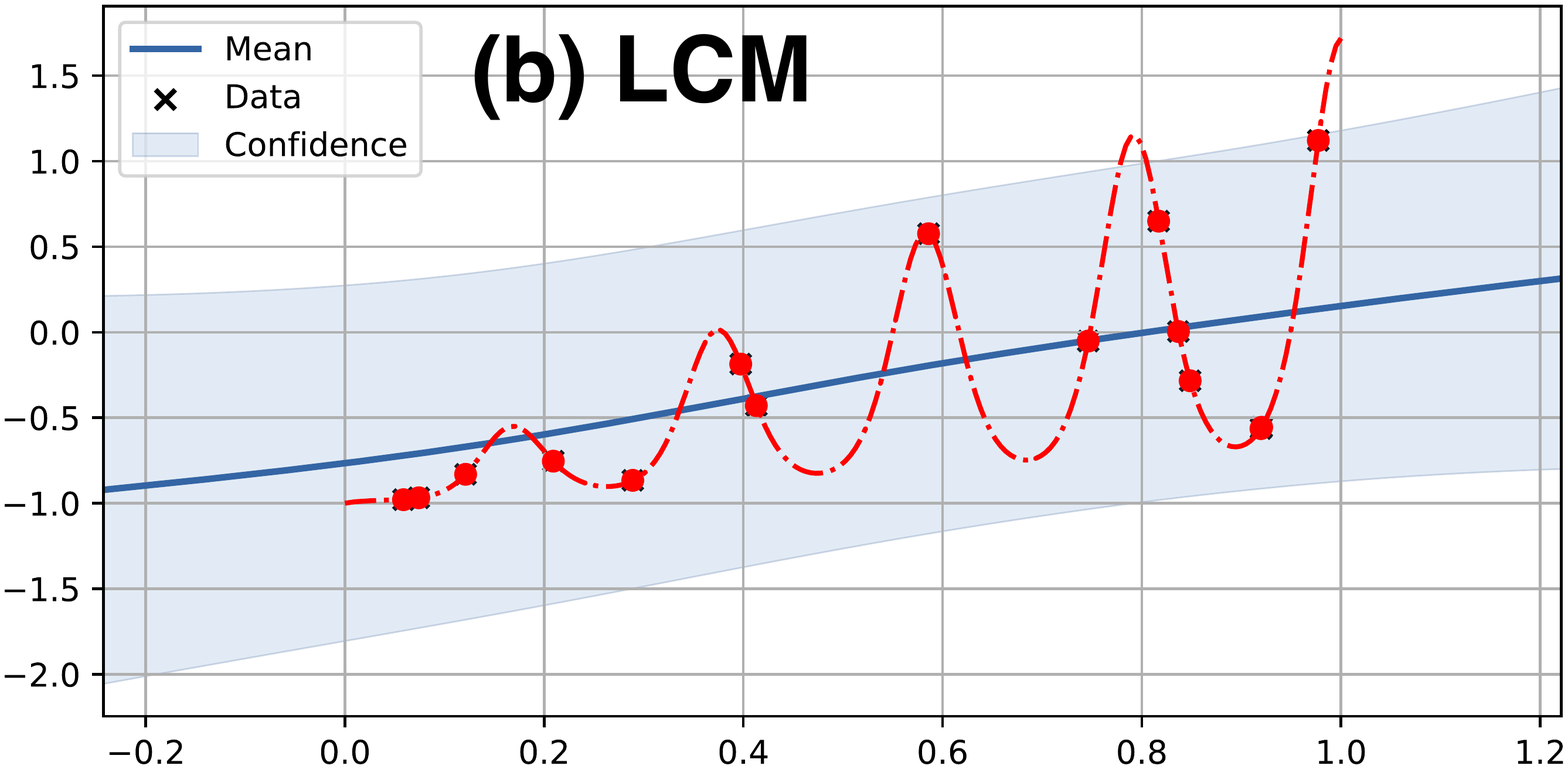}
  \includegraphics[width=0.3\columnwidth]{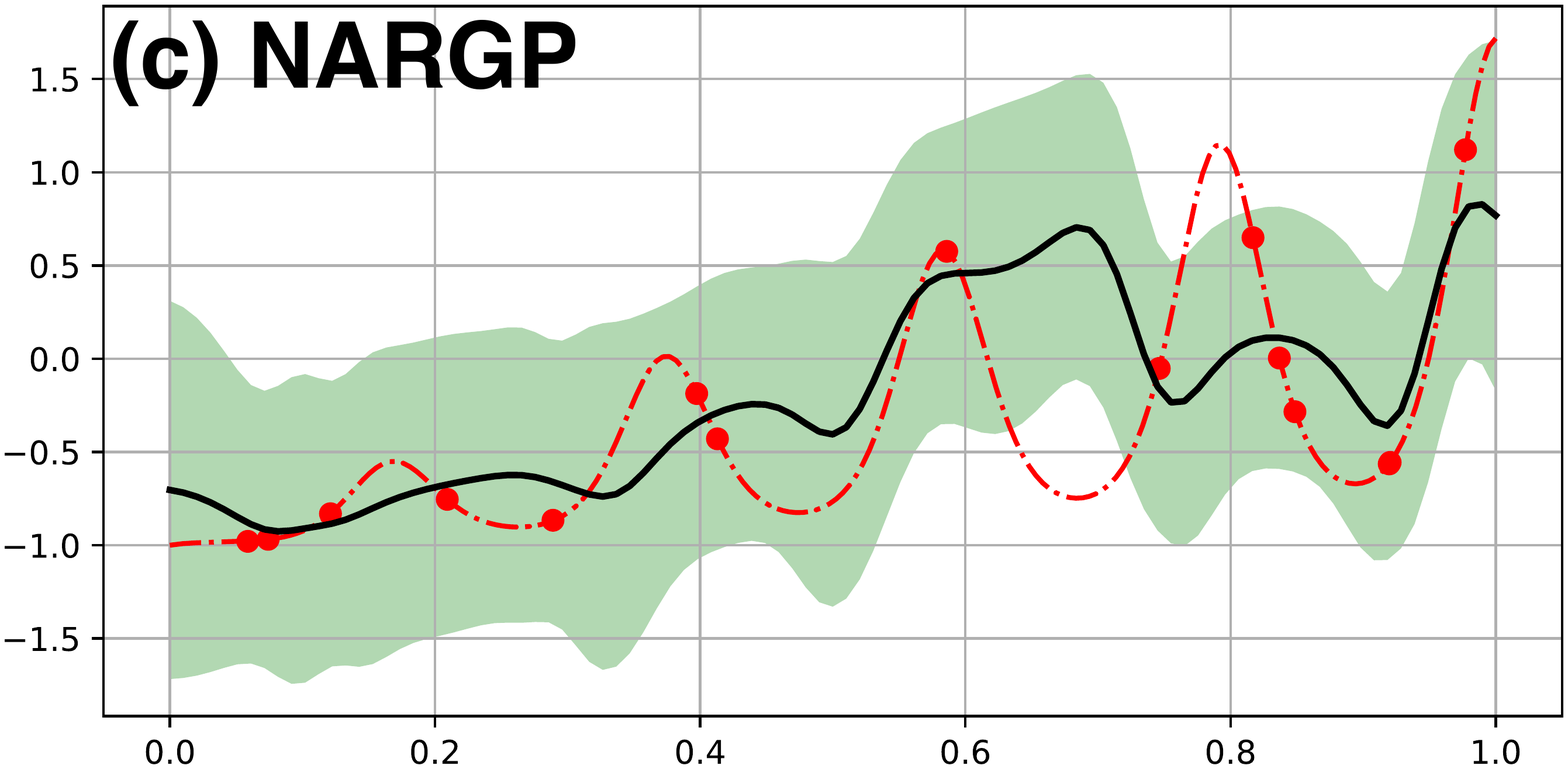}
  \includegraphics[width=0.3\columnwidth]{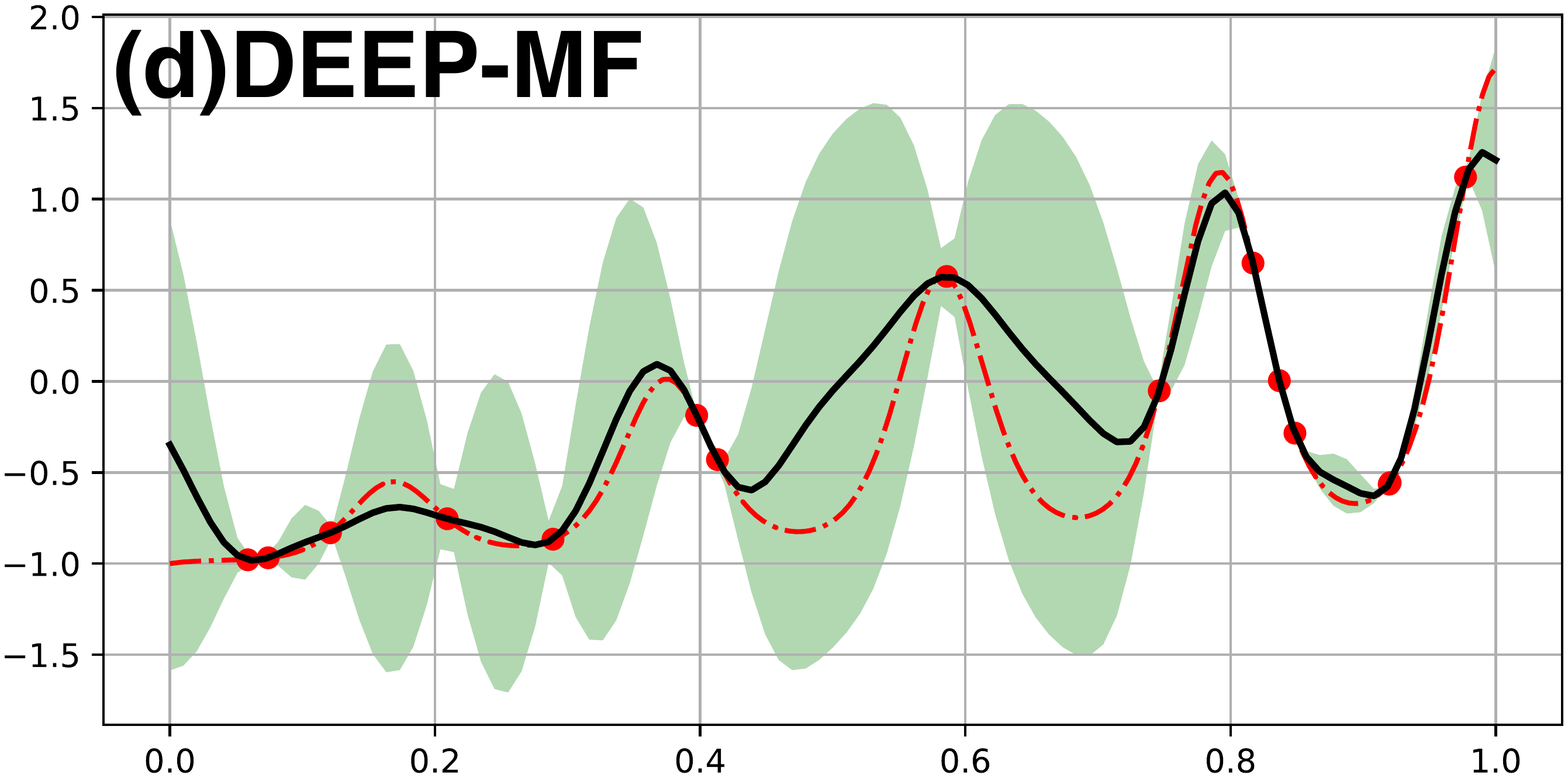}
  \includegraphics[width=0.3\columnwidth]{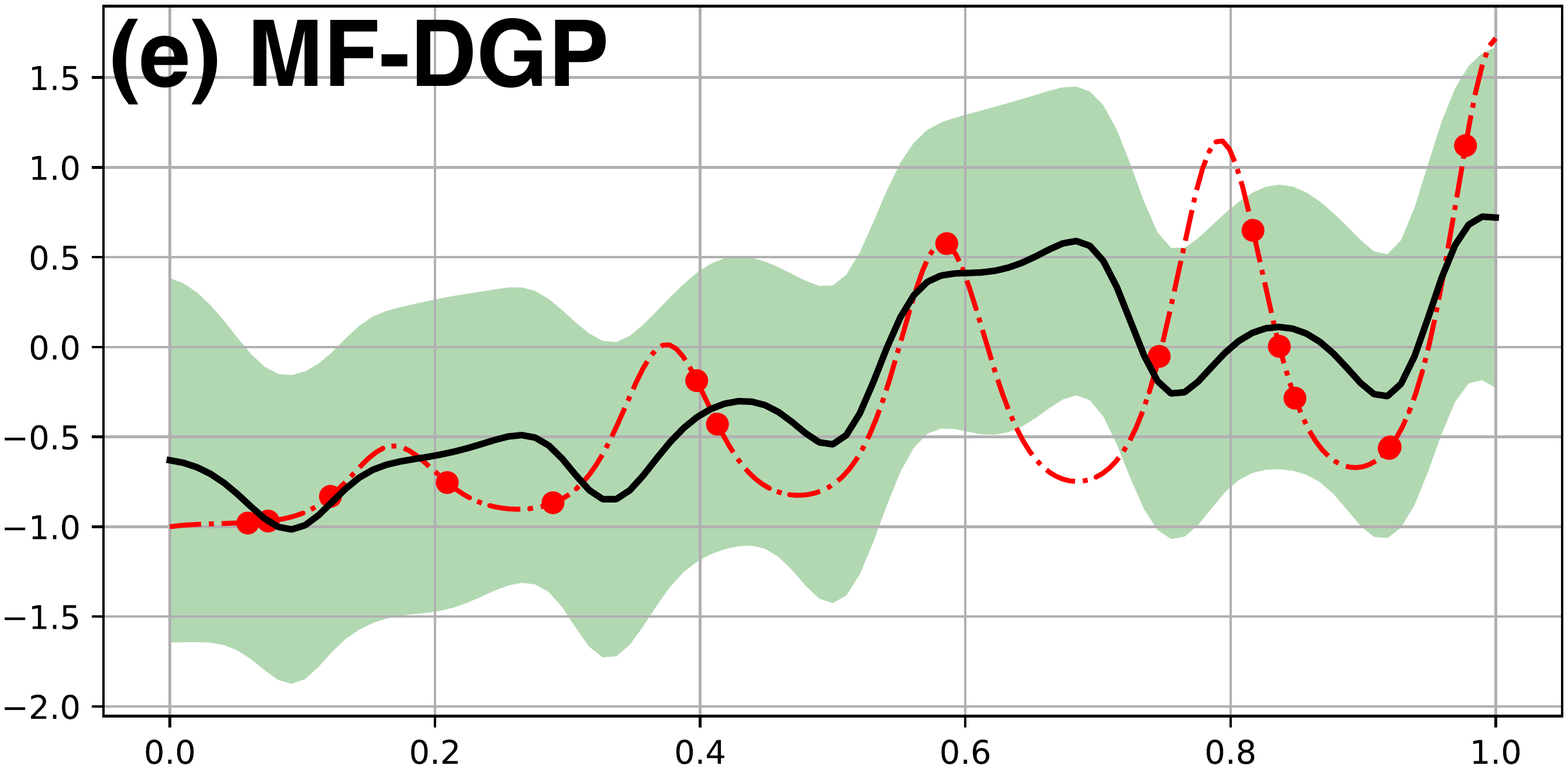}
  \includegraphics[width=0.3\columnwidth]{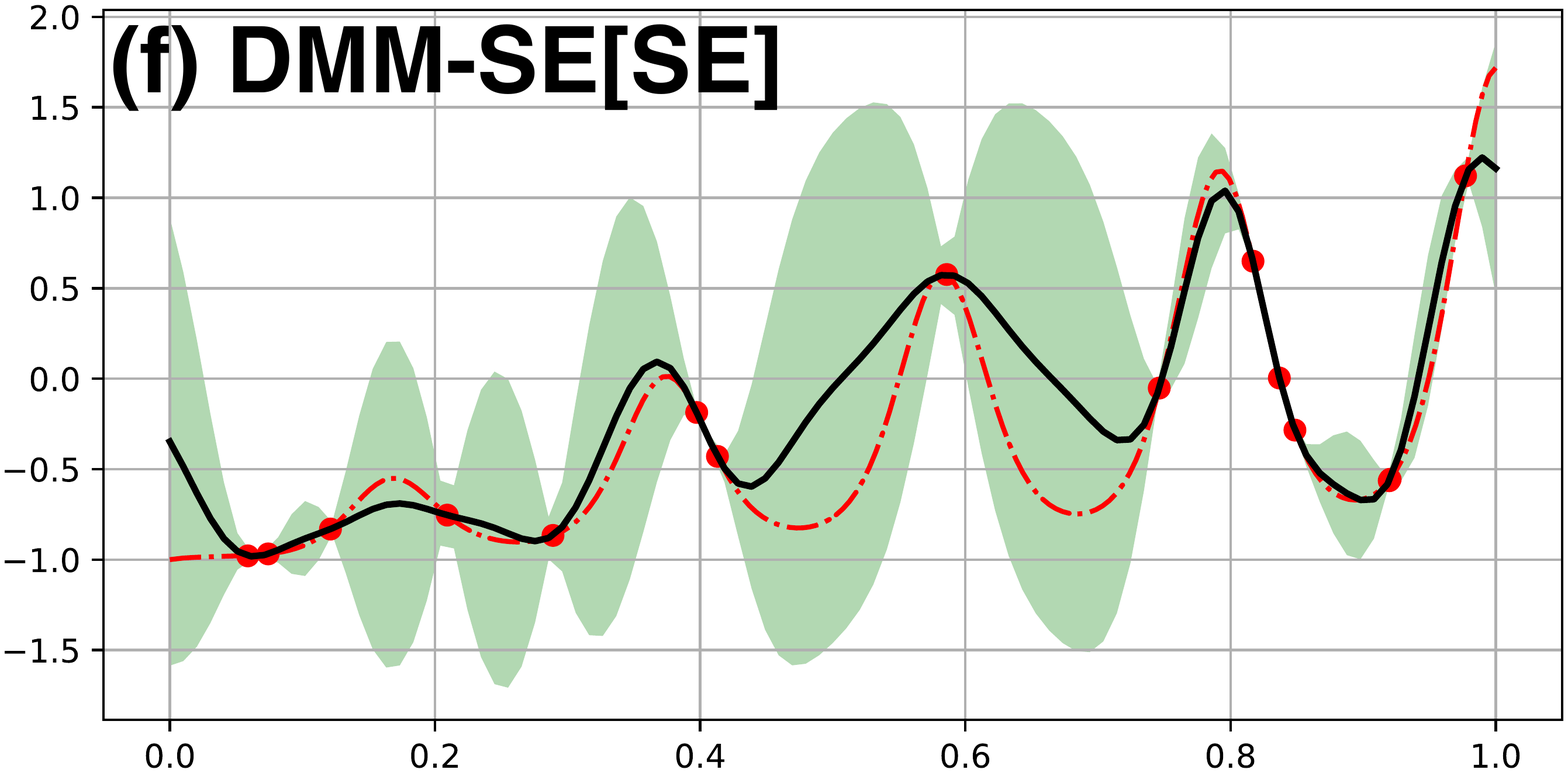}
  \caption{Multi-fidelity regression on the low-level true function, $h(x)=\cos15x$, with 30 observations and high-level one, $f(x)=x\exp[h(2x-0.2)]-1$, with 15 observations. (top row: (a) AR1, (b) LCM, and (c) NARGP. bottom row: (d) DEEP-MF, (e) MF-DGP, and (f) DMM-SE[SE])
  }
  \label{exp_nonlinearB}
\end{figure}

Figure~\ref{exp_nonlinearB} demonstrates another example of multi-fidelity regression on the nonlinear composite function. The low fidelity function is also periodic, $f_1=\cos15x$, and the target is exponential function, $f=x\exp[f_1(2x-2)]-1$. The 15 observations of $f$ (red dashed line) are marked by the red dots. The exponential nature in the mapping $f_1\mapsto f$ might make the reconstruction more challenging than the previous case, which may lead to less satisfying result from LCM [panel (b)]. NARGP [panel (c)] and MF-DGP [panel (e)] have similar predictions which mismatch some of the observations, but the target function is mostly covered by the uncertainty estimation. Our model with SE[SE] kernel [panel (f)], on the other hand, has predictions consistent with all the target observations, and the target function is fully covered by the uncertainty region. Qualitatively similar results are also obtained from AR1 [panel (a)] and DEEP-MF [panel (d)]. 

\subsection{Compositional freedom and varying low-fidelity data}

Given the good learning results in the previous subsection, one may wonder the effects of having a different low fidelity data set on inferring the high fidelity function. Here, we consider the same high fidelity data from the target function in Fig.~\ref{exp_nonlinearA}, but the low fidelity data are observations of $f_1(x)=x$, $f_1(x)=\tanh x$, $f_1(x)=\sin4\pi x$, and $f_1(x)=\sin8\pi x$. Plots in the top row represent $f_1|{\bf X}_1,{\bf y}_1$, while the bottom row shows the inferred target function given the high fidelity data (red dots). It can be seen in the left most column in panel (a) that the linear $f_1$ is not a probable low fidelity function as the true target function (red dashed line) in bottom is outside the predictive confidence. Similarly in the second plot in (a), $f_1$ being a hyper tangent function is not probable to account for the true target function. 
In the end, $f_1$ being a periodic function is more likely to account for the true target function than the first two cases, but the right most plot with $f_1(x)=\sin8\pi x$ leads to the the predictive mean very close to the true target function. 

Next, the low fidelity data become the noisy observations of the same four functions. As shown in panel (b), the increased variance in $f_1|{\bf X}_1,{\bf y}_1$ also results in raising the variance in $f$, especially comparing the first two cases in (a) against those in (b). A dramatic difference can be found in comparing the third plot in (a) against that in (b). In (b), the presence of noise in the low fidelity data slightly raises the uncertainty in $f_1$, but the ensuing inference in $f$ generates the prediction which fails to contain the most of the true target function within its model confidence. Thus, the likelihood that $f_1(x)=\sin4\pi x$ is the probable low fidelity function is greatly reduced by the noise in the observation. Lastly, the noise in observing $f_1(x)=\sin8\pi x$ as the low fidelity data does not significantly change the inferred target function. 

\begin{figure}[H]
\widefigure
    \centering
    \begin{subfigure}{6.5 cm}
        \includegraphics[width=\textwidth]{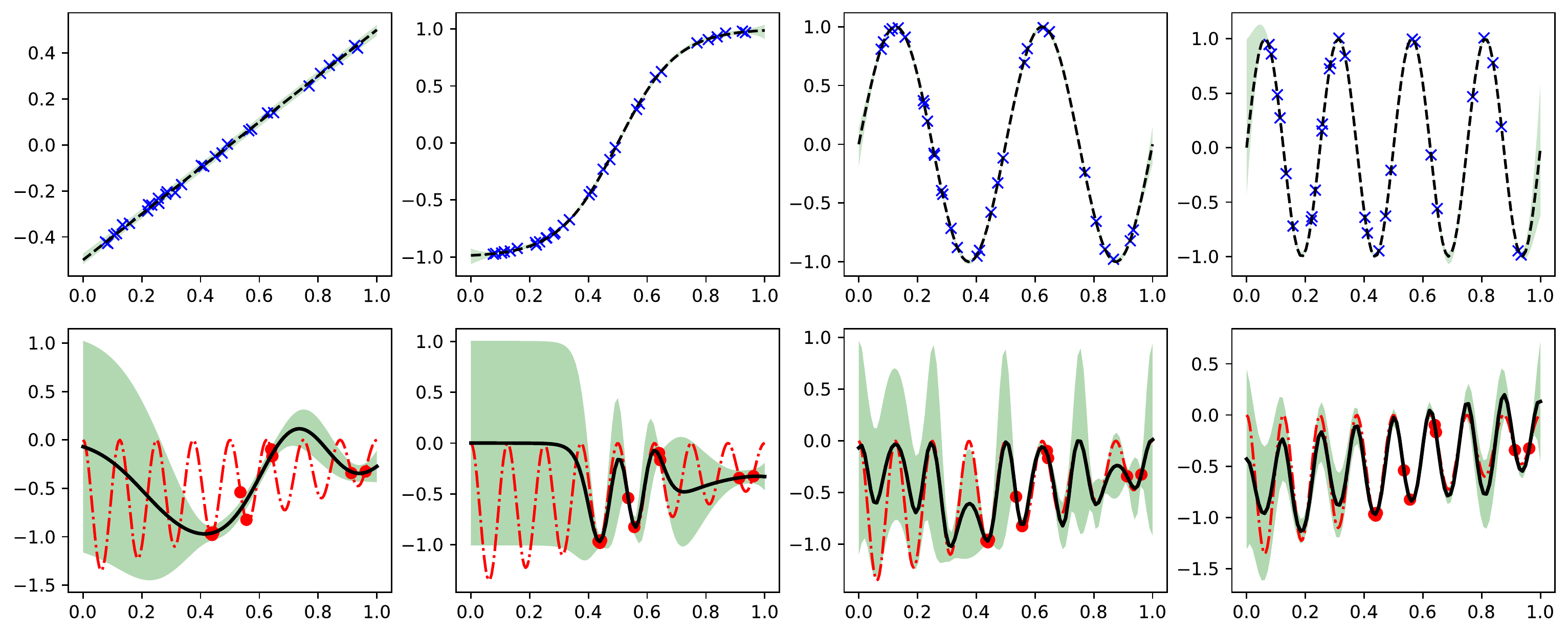}
        \caption{Low-fidelity data from 4 functions}
        \label{dgp1}
    \end{subfigure}
    ~
    \begin{subfigure}{6.5 cm}
        \includegraphics[width=\textwidth]{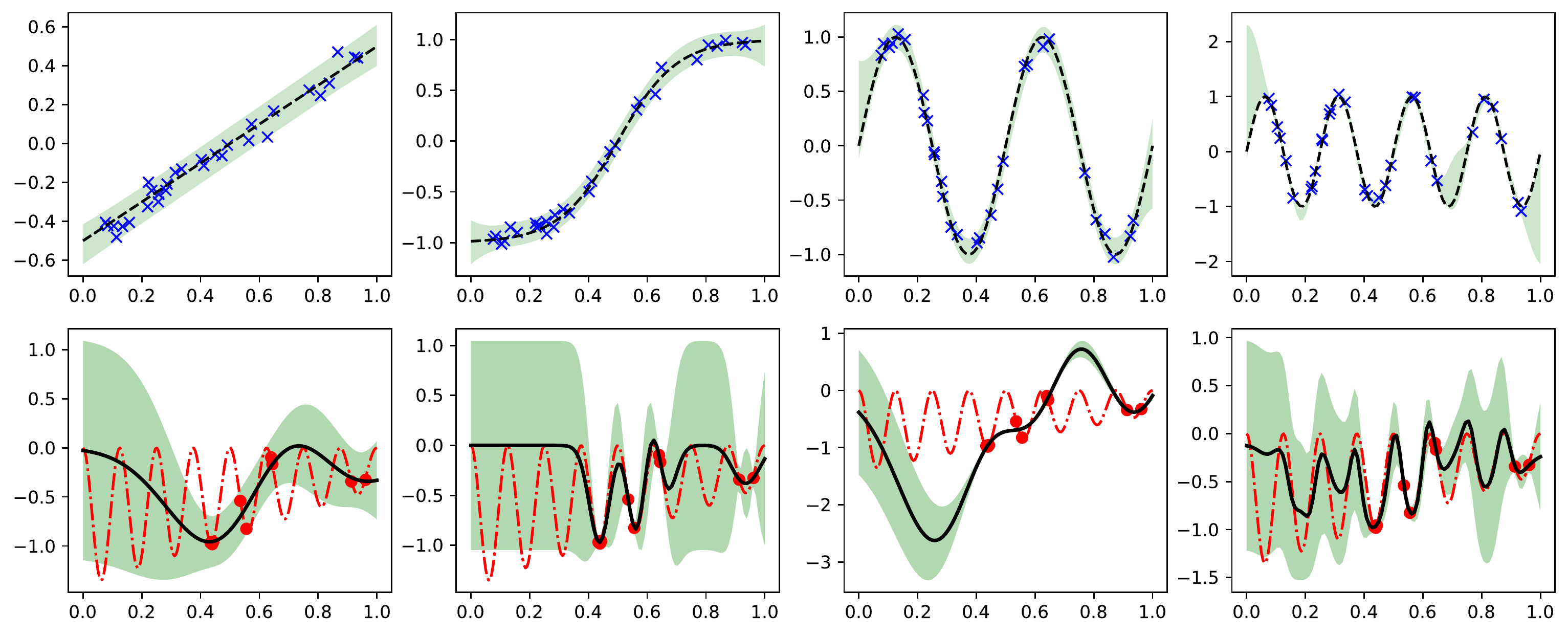}
        \caption{Low-fidelity data from noisy observations}
        \label{dgp2}
    \end{subfigure}
    \caption{Demonstration of compositional freedom and effects of uncertainty in low fidelity function $f_1$ on the target function inference. Given the same high fidelity observations of target function, four different sets of observations of $f_1(x)=x$, $f_1(x)=\tanh x$, $f_1(x)=\sin4\pi x$, and $f_1(x)=\sin8\pi x$ are employed as low fidelity data in inferring the target function. In panel (a), the low fidelity data are noiseless observations of the four functions. The true target function is partially outside the model confidence for the first two cases. In panel (b), the low fidelity data are noisy observations of the same four functions. Now the first three cases result in the inferred function outside the model confidence. The effect of uncertainty in low fidelity is most dramatic when comparing the third subplot in (a) and (b).}\label{fig:mm}
\end{figure}

\subsection{Denoising regression}

\begin{figure}[ht]
  \centering
  \includegraphics[width=0.6\columnwidth]{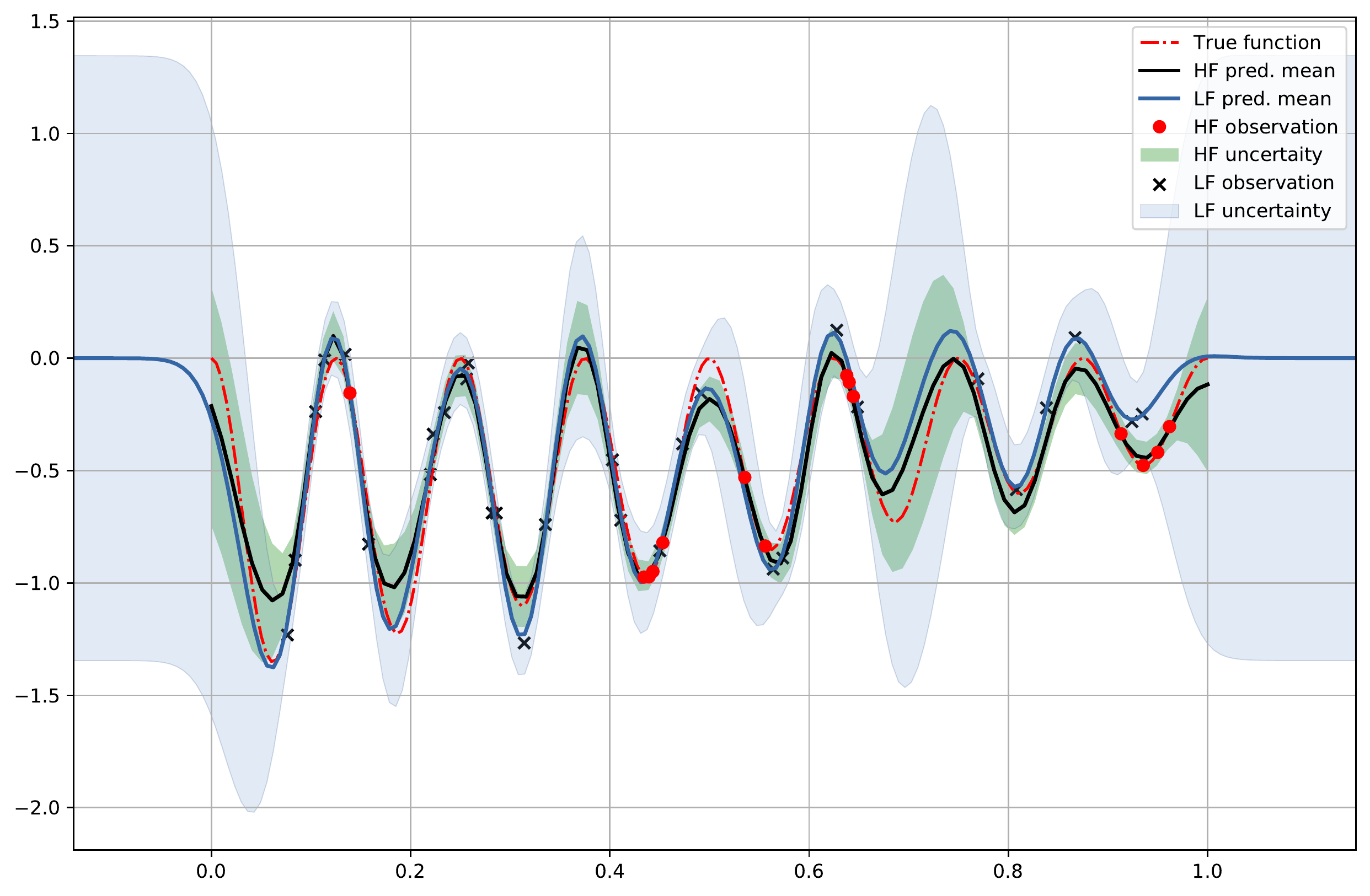}
  \caption{Denoising regression with 30 high-noise and 15 low-noise observations from the target function $y=(x-\sqrt{2})\sin^28\pi x$ (red dashed line). The uncertainty is reduced in the GP learning with the SE[SE] kernels.
  }
  \label{exp_syn_denoise}
\end{figure}

Here we continue considering the inference of the same target function in $f(x)=(x-\sqrt{2})\sin^28\pi x$,
but now the low fidelity data set becomes the noisy observations of the target function. See Fig.~\ref{exp_syn_denoise} for illustration. Now we have 
15 observations of $f$ with noise level of 0.001 (red dots) as high fidelity data and 
30 observations of the same function with nosie level of 0.1 (dark cross symbol) as the low fidelity data. Next, we follow the same procedure in inferring $f_1$ with the low fidelity, and then use the conditional mean and covariance in constructing the effective kernel for inferring the target function $f$ with the high fidelity data. Unlike the previous cases, the relation between $f$ and $f_1$ here is not clear.
However, the structure of DGP can be viewed as the intermediate GP emitting infinitely many samples of $f_1$ into the exposed GP. Qualitatively, one can expect that the actual prediction for $f$ is the average over the GP models with different warping $f_1$. Consequently, we may expect the variance in predicting $f$ is reduced.  

Indeed, as shown in Figure~\ref{exp_syn_denoise}, the predictive variance using a GP with the low fidelity (high noise) observations only is marked by the light-blue region around the predictive mean (light-blue solid line). When the statistical information in $f_1|{\bf X}_1,{\bf y}_1$ is transferred to the effective kernel, the new prediction and model confidence possess much tighter uncertainty (marked by the light-green shaded region) around the improved predictive mean (dark solid line) even in the region away from the low-noise observations.

\subsection{Multi-fidelity data with high dimensional input}

The work in~\citep{cutajar2019deep} along with their public code in emukit assembles a set of multi-fidelity regression data sets in which the input $\bf x$ is of high dimension. Here we demonstrate the simulation results on these data (see \cite{cutajar2019deep} for details). 
The simulation is performed using the effective kernels with compositions: SE[SE] and SC[SE] for the Borehole (two-fidelity) regression data set, SE[SE[SE]]] and SC[SC[SE]] for Branin (three-fidelity) regression data set. Algorithm~\ref{algorithm} is followed to obtain the results here. The performance of generalization is measured in terms of mean negative log likelihood (MNLL). Table~\ref{MFtable} displays the results using the same random seed from MF-DGP and our methods. We also include the simulation of standard GP regression with the high fidelity data only. It is seen that the knowledge about the low fidelity function is significant for predicting high-level simulation (comparing with vanilla GP) and that the informative kernels have better performance in these cases. 

\begin{table}[ht]
\caption{MNLL results of multi-fidelity regression.} 
\begin{center}
\begin{tabular}{lllll}
& \text{MFDGP}       & \text{SE[ ]}       & \text{SC[ ]}       & \text{GP+$\mathcal D_f$}     \\
\hline \\
Borehole   &  -1.87 &  ${\bf -2.08}$ & ${\bf -2.08}$ & 0.56 \\
Branin   & -2.7 &  -2.52 & ${\bf -2.93}$ & 5180 \\
\end{tabular}\label{MFtable}
\end{center}
\end{table}



\section{Discussion}


In this paper we propose a novel kernel learning which is able to fuse data of low fidelity into a prior for high fidelity function. 
Our approach addresses two limitations of prior research on GPs: the need to choose or design kernel~\cite{duvenaud2013structure,sun2018differentiable} and the lack of explicit dependence on the observations in the prediction (in Student-t process~\cite{shah2014student} the latter is possible). We resolve limitations associated with reliance on designing kernels, introducing new data-dependent kernels together with effective approximate inference. Our results show that the method is effective, and we prove that our moment-matching approximation retains some of multi-scale, multi-frequency, and non-stationary correlation that are characteristic of deep kernels, e.g.~\cite{wilson2016deep}. The compositional freedom~\cite{ustyuzhaninov2020compositional} pertaining to a hierarchical learning is also manifested in our approach.

\section{Conclusions}
Central to the allure of Bayesian methods, including Gaussian Processes, is the ability to calibrate model uncertainty through marginalization over hidden variables. The power and promise of DGP is in allowing rich composition of functions while maintaining the Bayesian character of inference over unobserved functions. Modeling the multi-fidelity data with the hierarchical DGP is able to exploit its expressive power and to consider the effects of uncertainty propagation. Whereas most approaches are based on variational approximations for inference and Monte Carlo sampling in prediction stage, our approach uses a moment-based approximation in which the marginal prior of DGP is a analytically approximated with a GP. For both, the full implications of these approximations are unknown. 
Continued research is required to understand the full strengths and limitations of each approach. 





\vspace{6pt} 



\authorcontributions{
Conceptualization, C.-K.L. and P.S.; methodology, C.-K.L.; software, C.-K.L.; validation, C.-K.L. and P.S.; formal analysis, C.-K.L.; investigation, C.-K.L.; resources, C.-K.L.; data curation, C.-K.L.; writing---original draft preparation, C.-K.L.; writing---review and editing, C.-K.L. and P.S.; visualization, C.-K.L.; supervision, P.S.; project administration, C.-K.L.; funding acquisition, P.S. All authors have read and agreed to the published version of the manuscript.
}

\funding{
This research was funded by the Air Force Research Laboratory and DARPA under agreement number FA8750-17-2-0146.
}

\acknowledgments{We thank the helpful correspondences with the authors of~\cite{requeima2019gaussian}.}

\conflictsofinterest{The authors declare no conflict of interest. The funders had no role in the design of the study; in the collection, analyses, or interpretation of data; in the writing of the manuscript, or in the decision to publish the~results.} 



\abbreviations{Abbreviations}{
The following abbreviations are used in this manuscript:\\

\noindent 
\begin{tabular}{@{}ll}
DGP & Deep Gaussian Process\\
SE & Squared Exponential\\
SC & Squared Cosine\\
\end{tabular}}

\appendixtitles{no} 
\appendixstart
\appendix
\section{}

\begin{figure}[ht]
  \centering
  \includegraphics[width=0.9\columnwidth]{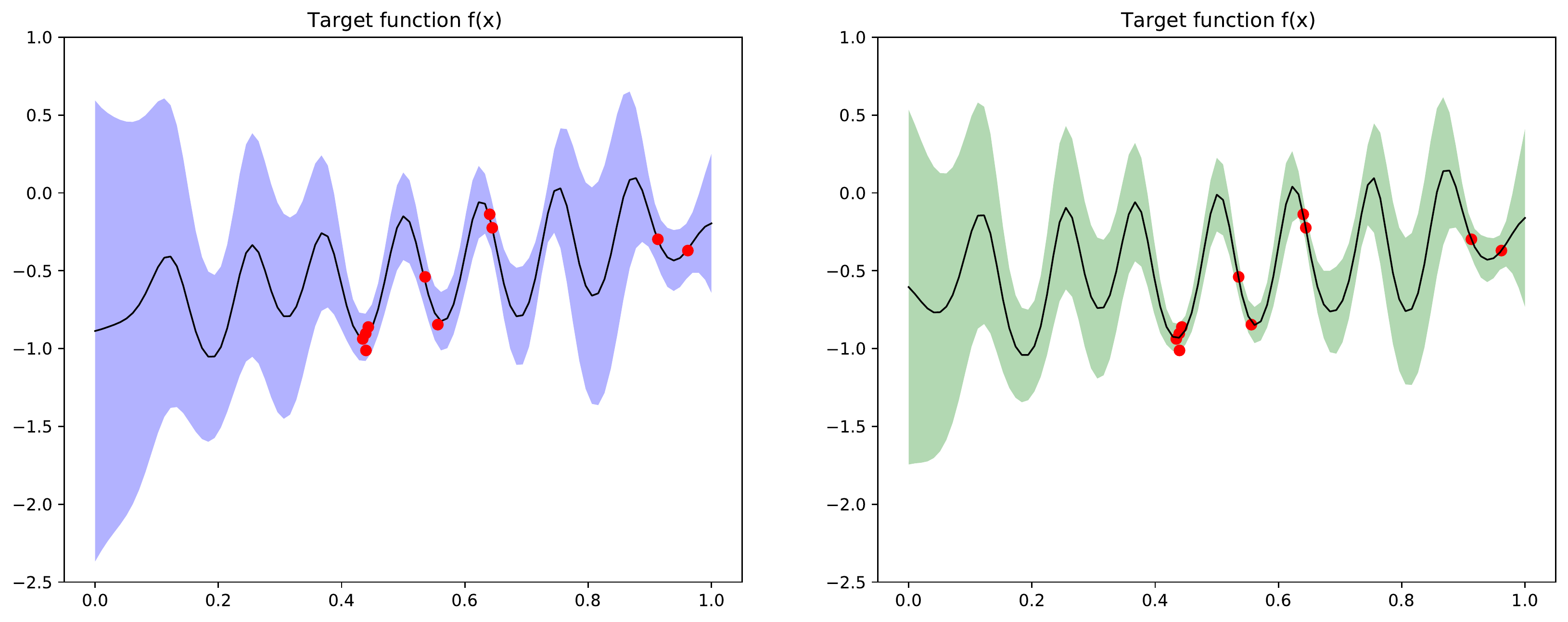}
  \caption{Comparison between the joint learning (left) and the sequential learning with Alg.~{\ref{algorithm}} (right). The same 10 training data are shown by the red dots. The joint learning algorithm results in a log marginal likelihood $1.65$ while the alternative one $2.64$. The hyper-parameters are $\{\sigma_{1,2}=(3.3,1.24), \ell_{1,2}=(0.12,1.40)\}$ (left) and $\{\sigma_{1,2}=(1.22,1.65), \ell_{1,2}=(0.08,0.98)\}$ (right).
  }
  \label{demo_synthetic_compare}
\end{figure}

Fig.~\ref{demo_synthetic_compare} shows the two results of multi-fidelity regressions with the same data. The left panel is obtained with jointly learning the hyperparameters in all layers with the standard gradient descent on the approximate marginal likelihood, while the right panel is from learning the hyperparameters sequentially with the Alg.~\ref{algorithm}. It is noted that the right panel yields higher log of marginal likelihood.




\end{paracol}

\reftitle{References}
\externalbibliography{yes}

\begin{thebibliography}{999}

\bibitem[Cutajar \em{et~al.}(2019)Cutajar, Pullin, Damianou, Lawrence, and
  Gonz{\'a}lez]{cutajar2019deep}
Cutajar, K.; Pullin, M.; Damianou, A.; Lawrence, N.; Gonz{\'a}lez, J.
\newblock Deep gaussian processes for multi-fidelity modeling.
\newblock {\em arXiv preprint arXiv:1903.07320} {\bf 2019}.

\bibitem[Lu \em{et~al.}(2020)Lu, Yang, Hao, and Shafto]{lu2020interpretable}
Lu, C.K.; Yang, S.C.H.; Hao, X.; Shafto, P.
\newblock Interpretable deep Gaussian processes with moments.
\newblock  International Conference on Artificial Intelligence and Statistics,
  2020, pp. 613--623.

\bibitem[Havasi \em{et~al.}(2018)Havasi, Hern{\'a}ndez-Lobato, and
  Murillo-Fuentes]{havasi2018inference}
Havasi, M.; Hern{\'a}ndez-Lobato, J.M.; Murillo-Fuentes, J.J.
\newblock Inference in deep Gaussian processes using stochastic gradient
  Hamiltonian Monte Carlo.
\newblock  Advances in Neural Information Processing Systems,  2018, pp.
  7506--7516.

\bibitem[Ustyuzhaninov \em{et~al.}(2020)Ustyuzhaninov, Kazlauskaite, Kaiser,
  Bodin, Campbell, and Ek]{ustyuzhaninov2020compositional}
Ustyuzhaninov, I.; Kazlauskaite, I.; Kaiser, M.; Bodin, E.; Campbell, N.; Ek,
  C.H.
\newblock Compositional uncertainty in deep Gaussian processes.
\newblock  Conference on Uncertainty in Artificial Intelligence. PMLR,  2020,
  pp. 480--489.

\bibitem[Kennedy and O'Hagan(2000)]{kennedy2000predicting}
Kennedy, M.C.; O'Hagan, A.
\newblock Predicting the output from a complex computer code when fast
  approximations are available.
\newblock {\em Biometrika} {\bf 2000}, {\em 87},~1--13.

\bibitem[Wang \em{et~al.}(2021)Wang, Xing, Kirby, and Zhe]{wang2021multi}
Wang, Z.; Xing, W.; Kirby, R.; Zhe, S.
\newblock Multi-Fidelity High-Order Gaussian Processes for Physical Simulation.
\newblock  International Conference on Artificial Intelligence and Statistics.
  PMLR,  2021, pp. 847--855.

\bibitem[Salakhutdinov \em{et~al.}(2012)Salakhutdinov, Tenenbaum, and
  Torralba]{salakhutdinov2012one}
Salakhutdinov, R.; Tenenbaum, J.; Torralba, A.
\newblock One-shot learning with a hierarchical nonparametric bayesian model.
\newblock  Proceedings of ICML Workshop on Unsupervised and Transfer Learning,
  2012, pp. 195--206.

\bibitem[Finn \em{et~al.}(2017)Finn, Abbeel, and Levine]{finn2017model}
Finn, C.; Abbeel, P.; Levine, S.
\newblock Model-agnostic meta-learning for fast adaptation of deep networks.
\newblock  International Conference on Machine Learning. PMLR,  2017, pp.
  1126--1135.

\bibitem[Titsias \em{et~al.}(2019)Titsias, Schwarz, Matthews, Pascanu, and
  Teh]{titsias2019functional}
Titsias, M.K.; Schwarz, J.; Matthews, A.G.d.G.; Pascanu, R.; Teh, Y.W.
\newblock Functional Regularisation for Continual Learning with Gaussian
  Processes.
\newblock  International Conference on Learning Representations,  2019.

\bibitem[Rasmussen and Williams(2006)]{rasmussen2006gaussian}
Rasmussen, C.E.; Williams, C.K.I.
\newblock {\em Gaussian Process for Machine Learning}; MIT press: Cambridge,
  MA,  2006.

\bibitem[Duvenaud \em{et~al.}(2013)Duvenaud, Lloyd, Grosse, Tenenbaum, and
  Zoubin]{duvenaud2013structure}
Duvenaud, D.; Lloyd, J.; Grosse, R.; Tenenbaum, J.; Zoubin, G.
\newblock Structure discovery in nonparametric regression through compositional
  kernel search.
\newblock  International Conference on Machine Learning. PMLR,  2013, pp.
  1166--1174.

\bibitem[Sun \em{et~al.}(2018)Sun, Zhang, Wang, Zeng, Li, and
  Grosse]{sun2018differentiable}
Sun, S.; Zhang, G.; Wang, C.; Zeng, W.; Li, J.; Grosse, R.
\newblock Differentiable compositional kernel learning for Gaussian processes.
\newblock  International Conference on Machine Learning. PMLR,  2018, pp.
  4828--4837.

\bibitem[Damianou and Lawrence(2013)]{damianou2013deep}
Damianou, A.; Lawrence, N.
\newblock Deep gaussian processes.
\newblock  Artificial Intelligence and Statistics,  2013, pp. 207--215.

\bibitem[Snelson \em{et~al.}(2004)Snelson, Ghahramani, and
  Rasmussen]{snelson2004warped}
Snelson, E.; Ghahramani, Z.; Rasmussen, C.E.
\newblock Warped gaussian processes.
\newblock  Advances in neural information processing systems,  2004, pp.
  337--344.

\bibitem[L{\'a}zaro-Gredilla(2012)]{lazaro2012bayesian}
L{\'a}zaro-Gredilla, M.
\newblock Bayesian warped Gaussian processes.
\newblock  Advances in Neural Information Processing Systems,  2012, pp.
  1619--1627.

\bibitem[Salimbeni and Deisenroth(2017)]{salimbeni2017doubly}
Salimbeni, H.; Deisenroth, M.
\newblock Doubly stochastic variational inference for deep gaussian processes.
\newblock  Advances in Neural Information Processing Systems,  2017.

\bibitem[Salimbeni \em{et~al.}(2019)Salimbeni, Dutordoir, Hensman, and
  Deisenroth]{salimbeni2019deep}
Salimbeni, H.; Dutordoir, V.; Hensman, J.; Deisenroth, M.P.
\newblock Deep Gaussian Processes with Importance-Weighted Variational
  Inference.
\newblock {\em arXiv preprint arXiv:1905.05435} {\bf 2019}.

\bibitem[Yu \em{et~al.}(2019)Yu, Chen, Low, Jaillet, and
  Dai]{haibin2019implicit}
Yu, H.; Chen, Y.; Low, B.K.H.; Jaillet, P.; Dai, Z.
\newblock Implicit Posterior Variational Inference for Deep Gaussian Processes.
\newblock  Advances in Neural Information Processing Systems,  2019, pp.
  14502--14513.

\bibitem[Le~Gratiet and Garnier(2014)]{le2014recursive}
Le~Gratiet, L.; Garnier, J.
\newblock Recursive co-kriging model for design of computer experiments with
  multiple levels of fidelity.
\newblock {\em International Journal for Uncertainty Quantification} {\bf
  2014}, {\em 4}.

\bibitem[Raissi and Karniadakis(2016)]{raissi2016deep}
Raissi, M.; Karniadakis, G.
\newblock Deep multi-fidelity Gaussian processes.
\newblock {\em arXiv preprint arXiv:1604.07484} {\bf 2016}.

\bibitem[Perdikaris \em{et~al.}(2017)Perdikaris, Raissi, Damianou, Lawrence,
  and Karniadakis]{perdikaris2017nonlinear}
Perdikaris, P.; Raissi, M.; Damianou, A.; Lawrence, N.; Karniadakis, G.E.
\newblock Nonlinear information fusion algorithms for data-efficient
  multi-fidelity modelling.
\newblock {\em Proceedings of the Royal Society A: Mathematical, Physical and
  Engineering Sciences} {\bf 2017}, {\em 473},~20160751.

\bibitem[Requeima \em{et~al.}(2019)Requeima, Tebbutt, Bruinsma, and
  Turner]{requeima2019gaussian}
Requeima, J.; Tebbutt, W.; Bruinsma, W.; Turner, R.E.
\newblock The gaussian process autoregressive regression model (gpar).
\newblock  The 22nd International Conference on Artificial Intelligence and
  Statistics. PMLR,  2019, pp. 1860--1869.

\bibitem[Alvarez \em{et~al.}(2011)Alvarez, Rosasco, and
  Lawrence]{alvarez2011kernels}
Alvarez, M.A.; Rosasco, L.; Lawrence, N.D.
\newblock Kernels for vector-valued functions: A review.
\newblock {\em arXiv preprint arXiv:1106.6251} {\bf 2011}.

\bibitem[Parra and Tobar(2017)]{parra2017spectral}
Parra, G.; Tobar, F.
\newblock Spectral mixture kernels for multi-output Gaussian processes.
\newblock  Proceedings of the 31st International Conference on Neural
  Information Processing Systems,  2017, pp. 6684--6693.

\bibitem[Bruinsma \em{et~al.}(2020)Bruinsma, Perim, Tebbutt, Hosking, Solin,
  and Turner]{bruinsma2020scalable}
Bruinsma, W.; Perim, E.; Tebbutt, W.; Hosking, S.; Solin, A.; Turner, R.
\newblock Scalable Exact Inference in Multi-Output Gaussian Processes.
\newblock  International Conference on Machine Learning. PMLR,  2020, pp.
  1190--1201.

\bibitem[Kaiser \em{et~al.}(2018)Kaiser, Otte, Runkler, and
  Ek]{kaiser2018bayesian}
Kaiser, M.; Otte, C.; Runkler, T.; Ek, C.H.
\newblock Bayesian alignments of warped multi-output Gaussian processes.
\newblock  Advances in Neural Information Processing Systems,  2018, pp.
  6995--7004.

\bibitem[Kazlauskaite \em{et~al.}(2019)Kazlauskaite, Ek, and
  Campbell]{kazlauskaite2019gaussian}
Kazlauskaite, I.; Ek, C.H.; Campbell, N.
\newblock Gaussian Process Latent Variable Alignment Learning.
\newblock  The 22nd International Conference on Artificial Intelligence and
  Statistics,  2019, pp. 748--757.

\bibitem[Williams(1997)]{williams1997computing}
Williams, C.K.
\newblock Computing with infinite networks.
\newblock  Advances in neural information processing systems,  1997, pp.
  295--301.

\bibitem[Cho and Saul(2009)]{cho2009kernel}
Cho, Y.; Saul, L.K.
\newblock Kernel methods for deep learning.
\newblock  Advances in neural information processing systems,  2009, pp.
  342--350.

\bibitem[Duvenaud \em{et~al.}(2014)Duvenaud, Rippel, Adams, and
  Ghahramani]{duvenaud2014avoiding}
Duvenaud, D.; Rippel, O.; Adams, R.; Ghahramani, Z.
\newblock Avoiding pathologies in very deep networks.
\newblock  Artificial Intelligence and Statistics,  2014, pp. 202--210.

\bibitem[Dunlop \em{et~al.}(2018)Dunlop, Girolami, Stuart, and
  Teckentrup]{dunlop2018deep}
Dunlop, M.M.; Girolami, M.A.; Stuart, A.M.; Teckentrup, A.L.
\newblock How deep are deep Gaussian processes?
\newblock {\em The Journal of Machine Learning Research} {\bf 2018}, {\em
  19},~2100--2145.

\bibitem[Shen \em{et~al.}(2020)Shen, Heinonen, and Kaski]{shen2020learning}
Shen, Z.; Heinonen, M.; Kaski, S.
\newblock Learning spectrograms with convolutional spectral kernels.
\newblock  International Conference on Artificial Intelligence and Statistics,
  2020, pp. 3826--3836.

\bibitem[Wilson \em{et~al.}(2016)Wilson, Hu, Salakhutdinov, and
  Xing]{wilson2016deep}
Wilson, A.G.; Hu, Z.; Salakhutdinov, R.; Xing, E.P.
\newblock Deep kernel learning.
\newblock  Artificial Intelligence and Statistics,  2016, pp. 370--378.

\bibitem[Daniely \em{et~al.}(2016)Daniely, Frostig, and
  Singer]{Daniely2016TowardDU}
Daniely, A.; Frostig, R.; Singer, Y.
\newblock Toward Deeper Understanding of Neural Networks: The Power of
  Initialization and a Dual View on Expressivity.
\newblock  NIPS,  2016.

\bibitem[Mairal \em{et~al.}(2014)Mairal, Koniusz, Harchaoui, and
  Schmid]{mairal2014convolutional}
Mairal, J.; Koniusz, P.; Harchaoui, Z.; Schmid, C.
\newblock Convolutional kernel networks.
\newblock  Advances in neural information processing systems,  2014, pp.
  2627--2635.

\bibitem[Van~der Wilk \em{et~al.}(2017)Van~der Wilk, Rasmussen, and
  Hensman]{van2017convolutional}
Van~der Wilk, M.; Rasmussen, C.E.; Hensman, J.
\newblock Convolutional gaussian processes.
\newblock  Advances in Neural Information Processing Systems,  2017, pp.
  2849--2858.

\bibitem[GPy(2012)]{gpy2014}
GPy.
\newblock {GPy}: A Gaussian process framework in python.
\newblock http://github.com/SheffieldML/GPy,  2012.

\bibitem[Wilson and Adams(2013)]{wilson2013gaussian}
Wilson, A.; Adams, R.
\newblock Gaussian process kernels for pattern discovery and extrapolation.
\newblock  International Conference on Machine Learning,  2013, pp. 1067--1075.

\bibitem[Shah \em{et~al.}(2014)Shah, Wilson, and Ghahramani]{shah2014student}
Shah, A.; Wilson, A.; Ghahramani, Z.
\newblock Student-t processes as alternatives to Gaussian processes.
\newblock  Artificial intelligence and statistics,  2014, pp. 877--885.

\end{thebibliography}
\end{document}